\documentclass{svproc}
\pdfoutput=1
\usepackage{epsfig}
\usepackage{calc}
\usepackage{amssymb}
\usepackage{amstext}
\usepackage{amsmath}
\usepackage{multicol}
\usepackage{pslatex}
\usepackage{subcaption}
\usepackage[noend]{algpseudocode}
\usepackage{caption}
\usepackage{algorithm}
\usepackage{graphicx}
\usepackage{csquotes}
\usepackage{url}
\usepackage[absolute,overlay]{textpos}

\begin{document}
\mainmatter

\title{On Randomized Searching for Multi-Robot Coordination}

\author{Jakub Hv\v{e}zda\inst{1,2} \and Miroslav Kulich\inst{1} \and Libor P\v{r}eu\v{c}il\inst{1}}
\tocauthor{Jakub Hv\v{e}zda, Miroslav Kulich, and Libor P\v{r}eu\v{c}il\inst{1}}

\institute{Czech Institute of Informatics, Robotics, and Cybernetics,
Czech Technical University in Prague, Prague, Czech Republic
\and
Department of Cybernetics, Faculty of Electrical Engineering, Czech Technical University in Prague, Czech Republic\\
\email{hvezdjak@fel.cvut.cz, kulich@cvut.cz, preucil@cvut.cz}\\
WWW home page: \texttt{http://imr.ciirc.cvut.cz}
}

\maketitle

\begin{textblock*}{\textwidth}(1in+\hoffset+\oddsidemargin,1cm) 
\centering
\small
In: Informatics in Control, Automation and Robotics. Cham: Springer, 2020. p. 364-383. Lecture Notes in Electrical Engineering. vol. 613. ISSN 1876-1100. ISBN 978-3-030-31992-2.
\end{textblock*}


\begin{abstract}

In this text, we propose a novel approach for solving the coordination of a fleet of mobile robots, which consists of finding a set of collision-free trajectories for individual robots in the fleet.
This problem is studied for several decades, and many approaches have been introduced.
However, only a small minority is applicable in practice because of their properties - small computational requirement, producing solutions near-optimum, and completeness.
The approach we present is based on a multi-robot variant of Rapidly Exploring Random Tree algorithm (RRT) for discrete environments and significantly improves its performance.
Although the solutions generated by the approach are slightly worse than one of the best state-of-the-art algorithms presented in~\cite{terMors2010}, it solves problems where ter Mors's algorithm fails.

\end{abstract}


\section{{Introduction}}
Recent advances in mobile robotics and increased deployment of robotic systems in many practical applications lead to intensive research of multi-robot systems. 
One of the most important problems is the coordination of trajectories of individual robots/agents in such systems: given starting and destination positions of the robots, we are interested in finding their trajectories that do not collide with each other, and the overall cost is minimized. 
An optimization criterion can be a sum of lengths of the individual trajectories or the time the last robot reaches its destination position.

\begin{figure}[t]
    \centering
     \includegraphics[width=\textwidth]{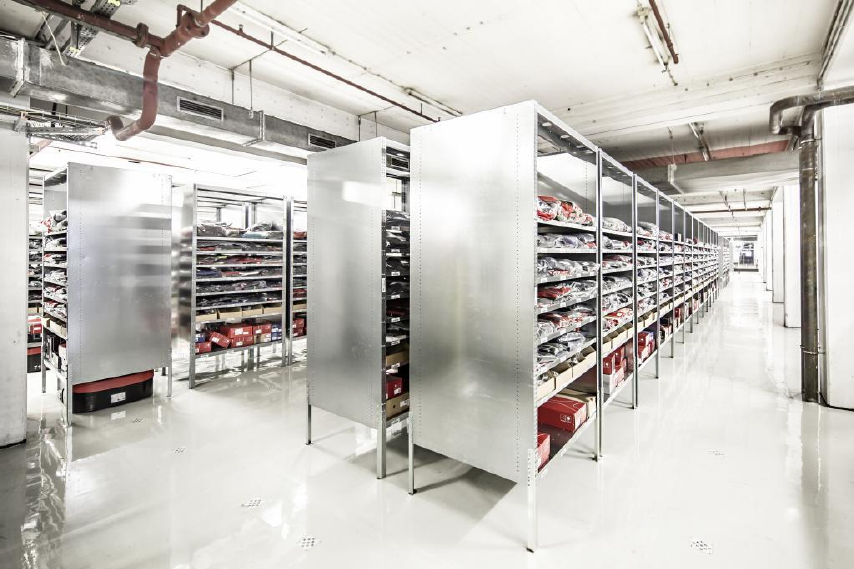}
      \caption{Automated warehouse: G-COM system by Grenzebach (\protect\url{https://www.grenzebach.com}) in a costumer application.}
    \label{fig:gcom}
\end{figure}

Several fields of the industry such as airports are nowadays faced with a higher increase in traffic than the actual capacity. 
This leads to reliance on path optimizations to increase their throughput.
Another typical application where multi-robot coordination plays an important role might be planning in an automated warehouse, see Fig.~\ref{fig:gcom}, where autonomous robots effectively deliver desired goods from/to given positions.
 
Multi-robot path planning and motion coordination has been studied since the 1980s, and many techniques have been developed during this period, see surveys~\cite{Parker2009,Doriya2015} for overview. This problem (formulated as the warehouseman's problem) was proved to be PSPACE-complete~\cite{Hopcroft1984}. 
For the case where robots move on a predefined graph, complexity of the problem can be reduced, nevertheless, it is still NP-hard~\cite{Goldreich11}, which means that optimal solutions cannot generally be found in a reasonable time for non-trivial instances (e.g., for a number of robots in order of tens).

Solutions to the problem consider either coupled or decoupled approaches. Centralized (coupled) approaches consider the multi-robot team as a multi-body robot for which classical single-robot path planning can be applied in composite configuration space. Traditional centralized methods are based on complete (i.e., the algorithm finds a solution if it exists or reports that no solution exists otherwise) and optimal classical algorithms and provide optimal solutions~\cite{Latombe1991},~\cite{Lavalle1998},~\cite{Ryan2008}. 
For example, a solution which assumes a multi-robot team as a multi-boty robot by building a Cartesian product of particular robots' configurations and finding a trajectory in the constructed space is presented in~\cite{Lavalle1998}.
However, these approaches require computational time exponential in the dimension of the composite configuration space, so they are appropriate for small-sized problems only. This drawback leads to the development of methods that prune the search space. For instance, Berg et al.~\cite{Berg2009} decompose any instance of the problem into a sequence of sub-problems where each subproblem can be solved independently from the others. The Biased Cost Pathfinding~\cite{Geramifard2006biased} employs generalized central decision maker that resolves collision points on paths that were pre-computed independently per unit, by replanning colliding units around the highest priority unit. Another approach is to design an algorithm based on a specific topology describing the environment. \cite{Peasgood08} present a multi-phase approach with linear time complexity based on searching a minimum spanning tree of the graph. 
The main idea of the algorithm is to find vertices for agents to move to while maintaining such a state of the
graph that does not block other agents. The paths between these vertices are then found using standard one-agent planning algorithms such as A* while looking at other agents as obstacles.

Flow Annotation Replanning, an approach for grid-like environments is introduced in~\cite{Wang2008}. A flow-annotated search graph inspired by two-way roads is built to avoid head-to-head collisions and to reduce the branching factor in an A* search. 
A heuristics is furthermore used to solve deadlocks locally instead of resorting to a more expensive replanning
step. Nevertheless, the computational complexity is still high (e.g.,~\cite{Berg2009} solves a problem with 40 robots in 12 minutes,~\cite{Wang2008} needs approx. 30 seconds for 400 robots).

On the contrary, decoupled methods present a coordination phase separated from the path planning phase. These approaches provide solutions typically in orders of magnitude faster times than coupled planners, but these solutions are sub-optimal. Moreover, the decoupled methods are often not complete as they may suffer from deadlocks. These approaches are divided into two categories: path coordination techniques and prioritized planning. Path coordination considers tuning the velocities of robots along the precomputed trajectories to avoid collisions like in~\cite{LaValle1998_OMP}.
Similarly, a resolution-complete algorithm is presented in~\cite{Simeon2002}, which is consists of exploring a so-called coordination diagram (CD). CD was firstly introduced in~\cite{pao89} and it describes conflicts of robots in a joint space where each dimension represents and individual robot's position on its path.

Prioritized planning computes trajectories sequentially for the particular robots based on the robots' priorities. Robots with already determined trajectories are considered as moving obstacles to be avoided by robots with lower priorities.
A simple heuristics to assign priorities is introduced in~\cite{VandenBerg2005} -- priority of a robot is determined as the distance to its goal.  
Another approach, a randomized hill-climbing technique based on a greedy local search procedure~\cite{Selman1992} to optimize the order of robots is presented in~\cite{Bennewitz2001}.
Finally, \v{C}\'ap~et~al.~\cite{Cap2015} discuss conflicts that occur during prioritized planning and propose a revised version of prioritized planning (RPP) that tries to avoid these conflicts. They also propose decentralized implementations of RPP (synchronous as well as asynchononous) and prove that the asynchronous version is guaranteed to terminate. 

Ter Morse et al.~\cite{terMors2010} present another prioritized planning scheme, but it codes information about trajectories of robots with higher priorities into a planning graph rather than into the planning algorithm itself. 
It does so by constructing a resource graph, where each resource can be for example a node of the original graph or intersection graph edges.
Every such resource holds information about time intervals in which it is not occupied by already planned robots.
An adaptation of the A* algorithm is used on this graph to find the shortest path through these intervals (called free time windows) to obtain a path that avoids all already planned robots.

Gawrilow et al.~\cite{Gawrilow2008} deal with a real-life problem of routing vehicles in Container Terminal Altenwerder in Hamburg harbor.
They used a similar approach to keeping a set of free time windows for path arcs in the graph.
Their algorithm contains a preprocessing of the graph for the use of specific vehicles followed by computation of paths for individual vehicles on this preprocessed graph.

Another similar approach is presented in \cite{spatioTempA*}, where each robot looks for a viable path in a 2D spatial grid map and checks for collisions with moving obstacles using a temporal occupancy table.
Zhang et al.~\cite{Zhang2015} adopt a similar approach to \cite{terMors2010} with enhanced taxiway modeling approach to improve performance on airport graph structures.
Ter Mors et al.~\cite{terMorsComparison} compare Context-Aware Route Planning~(CARP)~\cite{terMors2010} with a fixed-path scheduling algorithm using $k$ shortest paths~\cite{Yen1971} and a fixed-path scheduling algorithm using $k$ disjoint paths~\cite{Suurballe1974}.
The experiments show that the CARP algorithm is superior in all measured qualities.
In our recent paper~\cite{Hvezda2018ITSC}, we propose a modification of CARP, which generates a trajectory for
an robot $a_k$ assuming that trajectories for $k − 1$ robots are already planned which can possibly lead to modification of
those planned trajectories. The main idea is to iteratively build a set of robots whose trajectories mostly influence an optimal
trajectory of $a_k$. The experimental results show that the proposed approach finds better solutions than the original CARP algorithm after several random shuffles of the robots’ priorities while requiring significantly less computational time for adding individual robots into the system.

Comparison of several heuristic approaches of assigning priority to robots in~\cite{terMorsHeuristics} concludes that the heuristics which plans longest paths first perform best when a makespan is to be minimized. 
A greedy best-first heuristics provides best results regarding joint plan cost. However, its downside is that it calls the planning algorithm for all yet unplanned robots in every round and it is thus very time-consuming.

Several computationally efficient heuristics have been introduced recently enabling to solve problems for tens of robots in seconds. 
Chiew~\cite{Chiew2010} proposes an algorithm for $n^2$ vehicles on a $n\times n$ mesh topology of path network allowing simultaneous movement of vehicles in a corridor in opposite directions with computational complexity $O(n^2)$. 
Windowed Hierarchical Cooperative A* algorithm (WHCA*) employs heuristic search in a space-time domain based on hierarchical A* limited to a fixed depth~\cite{Silver2005}. 
Wang and Botea~\cite{Wang2011} identify classes of multi-robot path planning problems that can be solved in polynomial time and introduce an algorithm with low polynomial upper bounds for time, space and solution length. 
Luna and  Bekris~\cite{Luna2011} present a complete heuristics for general problems with at most $n-2$ robots in a graph with n vertices based on the combination of two primitives - \enquote{push} forces robots towards a specific path, while \enquote{swap} switches positions of two robots if they are to be colliding. An extension which divides the graph into subgraphs within which it is possible for agents to reach any position of the subgraph, and then uses \enquote{push}, \enquote{swap}, and \enquote{rotate} operations is presented in~\cite{DeWilde2014}.
Finally, Wang and Wooi~\cite{Wang2015} formulate multi-robot path planning as an optimization problem and approximate the objective function by adopting a maximum entropy function, which is minimized by a probabilistic iterative algorithm.

The approaches mentioned above have nice theoretical properties, but Context-Aware Route Planning (CARP)~\cite{terMors2010} is probably the most practically usable algorithm as it produces high quality solutions fast, and it finds a solution for a large number of practical setups.  

Recently, Solovey et al. present an approach for multi-robot coordination inspired by the Rapidly-exploring random tree (RRT) algorithm \cite{Lavalle1998}.
The approach, MRdRRT~\cite{drrt} is probabilistic and plans paths on predefined structures (graphs) for small fleets of robots.  
This is further improved in~\cite{Dobson2017} by employing ideas from RRT*\cite{Karaman2011}, a variant of RRT which converges towards an optimal solution.

In this paper, we present a probabilistic approach which extends and improves a discrete version of Rapidly-Exploring Random Tree (RRT) for multiple robots~\cite{drrt}.
Our approach focuses mainly on scalability with increasing number of agents as well as improving the quality of solution compared to \cite{Dobson2017} that presents the optimal version of the dRRT algorithm but keeps the number of robots relatively low.
We show that the proposed extensions allow solving problems with tens of robots in times comparable to CARP with a slightly worse quality of results. 
On the other hand, the proposed algorithm finds solutions also for setups where CARP fails.

The paper is an extended version of the paper~\cite{Hvezda2018ICINCO} presented at the 15th International Conference on Informatics in Control, Automation and Robotics (ICINCO). In comparison to the ICINCO paper, the following modifications were made:
\begin{itemize}
  \item The state-of the art was refined and extended and new references were added. 
  \item The section about the proposed improvements of the original MRdRRT algorithm were extended. Namely, the proposed use of the CARP algorithm was modified with a detailed decription of CARP.
  \item Experiments were performed on various sized of maps (only one map size was considered in the ICINCO paper).  Moreover, influence of two main improvements, expansion and rewiring, on solution quality and time complexity was studied.
  \item All the drawings were updated and redesigned and some additional figures were included.
\end{itemize}
The rest of the paper is organized as follows. 
The multi-agent path-finding problem is presented as well as the used terms are defined in Section~\ref{sec:problem}.
The multi-robot discrete RRT algorithm and the proposed improvements are described in Section~\ref{sec:algorithm}, while performed experiments, their evaluation, and discussion are presented in Section~\ref{sec:experiments}.
Finally, Section~\ref{sec:conclusion} is dedicated to concluding remarks.

\section{\uppercase{Problem definition}}
\label{sec:problem}

The problem of multi-agent pathfinding/coordination concerns itself with searching for non-colliding paths for each agent in a set of agents that starts in agents start location and ends in its goal location while minimizing some global cost function such as agent travel time or global plan completion time.

For more rigorous specification, we assume:
\begin{itemize}
\item A set of \textit{k} homogenous agents each labeled $a_1, a_2,..., a_k$.    
\item A graph $G(V,E)$ where $|V| = N$. The vertices $V$ of the graph are all possible agent's locations, while $E$ represents a set of all possible agent's transitions between the locations.
\item A start location $s_i \in V$ and a target location $t_i \in V$ of each agent.
\end{itemize}



The goal is finding a set of trajectories on $G(V,E)$ that avoid collisions with other robots and environmental obstacles that specify all locations of each agent in each time step such that the agents are in their initial positions at the start and in their goal positions at the end.

Key used terms and additional constraints to the generated trajectories will be specified in the following paragraphs.

\subsection{Actions}

In each time point, only two action types are permitted for every agent:
The first type of action is for the agent to move into one of the neighbouring nodes and the second type of action is that the robot waits at its current location.
The cost of these actions can be different for each algorithm, but in our case, the assumption is that staying still has zero cost. 
Also, another assumption is that once agents reach their goal location, they wait for all other agents to finish their respective plans.

\subsection{Constraints}
In this paper the main constraints placed upon the movement of agents are:

\begin{itemize}
\item No two agents $a_1$ and $a_2$ can occupy the same vertex $v \in V$ at the same time.
\item Assume two agents $a_1$, $a_2$ located in two neighboring nodes $v_1,v_2 \in V$ respectively, they can not travel along the same edge $(v_1,v_2)$ at the same time in opposite directions.
In other words, two neighboring agents cannot swap positions.
However, it is possible for agents to follow one another assuming that they do not share the same vertex or edge concurrently.
For example, if the agent $a_1$ moves from $v_2 \in V$ to $v_3 \in V$ then the agent $a_2$ can move from $v_1 \in V$ to $v_2 \in V$ at the same time.
\end{itemize}

\subsection{Composite configuration space}



The \textit{composite configuration space} $\mathcal{G} = (\mathcal{V},\mathcal{E})$ is a graph that can be defined in a following manner.
The vertices $\mathcal{V}$ are all combinations of collision-free placements of \textit{m} agents on the original graph $G$.
These vertices can also be viewed as $m$ agent configurations $C = (v_1,v_2,...,v_m)$, where an agent $a_i$ is located at a vertex $v_i \in G$ and the agents do not collide with each other.
The edges of $\mathcal{G}$ can be created using either Cartesian product or Tensor product.
In this paper the Tensor product is used because is allows multiple agents to move simultaneously. Therefore, for two $m$ agent configurations $C = (v_1,v_2,...,v_m)$, $C' = (v'_1,v'_2,...,v'_m)$ the edge $(C,C')$ exists if $(v_i,v'_i) \in E_i$ for every $i$ and no two agents collide with each other during the traversal of their respective edges.

The distance between two neighboring nodes $C_1=(v_{11},v_{12}, ... , v_{1n})$ and $C_2=(v_{21},$ $v_{22}, ... , v_{2n})$ in a composite roadmap is calculated as the sum of Euclidean distances $d$ between the corresponding nodes:

$$\delta\left(C_1,C_2\right) = \sum_{i = 0}^{n} d(v_{1i},v_{2i})$$
\section{\uppercase{Proposed algorithm}}
\label{sec:algorithm}
\subsection{Discrete RRT} \label{sec:drrt}

A discrete multi-robot rapidly-exploring random tree (MRdRRT)\cite{drrt} is a modification of the RRT algorithm for pathfinding in an implicitly given graph embedded in a high-dimensional Euclidean space.

Similarly to RRT, the MRdRRT finds the paths for the agents in the composite configuration space $\mathbb{R}^d$ by growing a tree $\mathcal{T}$ rooted in the vertex $s$ that represents the start locations of all agents.
The tree is expanded by iterative addition of new points while simultaneously attempting to connect the newly added points to the goal vertex $t$ such that no constraints are violated, e.g. not causing any collisions with the environment and between agents.
The addition of new points is handled first sampling a random point $u$ from the composite configuration space, followed by extending the current tree towards the sample point $u$.
The important thing to note here is that vertices added to the tree are taken from $G$.
This means that to extend the tree towards sample point it is required to first find the node in the tree from which the extension is made - in this case the nearest point $n$ in the tree to the sample $u$ is chosen, but also which neighbour of this node will be selected as the extension.
Because unlike in RRT that operates in continuous space, it is not possible to make a step in the given direction.
To choose the best neighbour of $n$ the MRdRRT uses a technique called oracle.
Without loss of generality consider that $G$ is embedded in $\left[0,1\right]^d$.
For two points $v,v' \in \left[0,1\right]^d$ the $\rho\left(v,v'\right)$ denotes a ray that begins in $v$ and goes through $v'$.
$\angle_v\left(v',v''\right)$ given three points $v,v',v'' \in \left[0,1\right]^d$ denotes the (smaller) angle between $\rho\left(v,v'\right)$ and $\rho\left(v,v''\right)$.
The way the oracle is used is given sample point $u$ it returns the neighbor $v'$ of $v$ such that the angle between rays $\rho\left(u,v'\right)$ and $\rho\left(v,v'\right)$ is minimized. This can be defined as

$$\mathcal{O}_D\left(v,u\right) := \underset{v' \in V}{\mathrm{argmin}} \left\lbrace \angle_v\left(u,v'\right)|\left(v,v'\right) \in E \right\rbrace.$$

As mentioned earlier the growth of the tree is not done only by adding single new nodes, but also attempting to connect the newly added nodes to the goal vertex $t$. The reason for this is that the tree may eventually expend to this vertex, but it is unlikely for larger problems.
For this reason, the {\em local connector} is necessary to ensure the algorithm finds the paths in a reasonable amount of time.
The main restriction for the {\em local connector} is that it returns decision about finding or not finding the paths quickly by attempting to solve only a restricted version of the problem so that it can be run often.

\subsection{Proposed improvements}

The original MRdRRT is able to solve path-finding problems for several agents.
However, the realisation of its particular steps is inefficient, which results in its inability to deal with complex scenarios in which tens of robots take part of.
The experimental results with up to 10 robots were presented by the authors of MRdRRT, with the mention that their algorithm faces problems once the tasks contain a more substantial number of robots.
To improve the behaviour of the algorithm we, therefore, introduced several modifications to the original version.
\subsubsection{Random sample generation improvement}

The version of the expansion phase proposed in the original paper generates random samples from the bounding box of $G$. However, this is inefficient in maps with tightly spaced because it disallows the agents to stay on their current position in the next step resulting in not being able to find a solution for problems where this action is necessary.
Also, most of the points that were generated by the original expansion phase procedure were far from a solution leading to a really unnecessarily large growth of the tree over the configuration space and resulting in high computational complexity of the algorithm.
Our solution to this issue is that we find shortest path for each agent separately during preprocessing and then create a set of all possible samples that can be generated for the particular agent as all points for which $dist(s_i, q) + dist(q, t_i) \leq dist(s_i, t_i) + \Delta$,  where $dist$ is a distance of two points, $s_i$ and $t_i$ are start and goal positions of $i$-th robot, and $\Delta>0$ is a defined constant threshold.
\subsubsection{Improvements to oracle method}

The main issue of the original oracle method was the fact that it checks for collisions after a new candidate point is generated. This results in many failed attempts and discarding several samples.
The version proposed here takes the collision possibility into account during the process of generating candidate point in an attempt to generate new points that are collision-free and thus reducing the number of iterations required to produce a viable new point. 
Just as the original version, our version also attempts to minimize the angle $(u_i,v_i,v_i')$.

\subsubsection{Proposed use of CARP algorithm as local connector}

The local connector in the original version of the algorithm works such that it finds the shortest paths for all robots not taking the other ones into account. It then attempts to find an order in which the robots would move to their destinations one by one while the others stand still. If such ordering exists, the local connector reports success.

We propose the use of CARP of algorithm~\cite{terMors2010} as a local connector along with random shuffling of the order in which CARP attempts planning of trajectories for individual agents.
CARP algorithm models the environment as a resource graph, that stores information about capacity and occupancy in time for each individual resource. 
For this reason, the resource graph can also be called {\em free time window graph}. 
CARP uses a modified A* algorithm on the time window graph to find the shortest path through time for each robot from his starting resource to his goal resource.
The found path takes the form of a sequence of time windows and corresponding resources, which it subsequently updated into the free time window graph to ensure the other agents can avoid the given agent.

\begin{figure*}
\label{fig:mors}	
    \begin{subfigure}[t]{0.50\linewidth}
   	\subcaption{CARP example problem}
        \includegraphics[width=\linewidth]{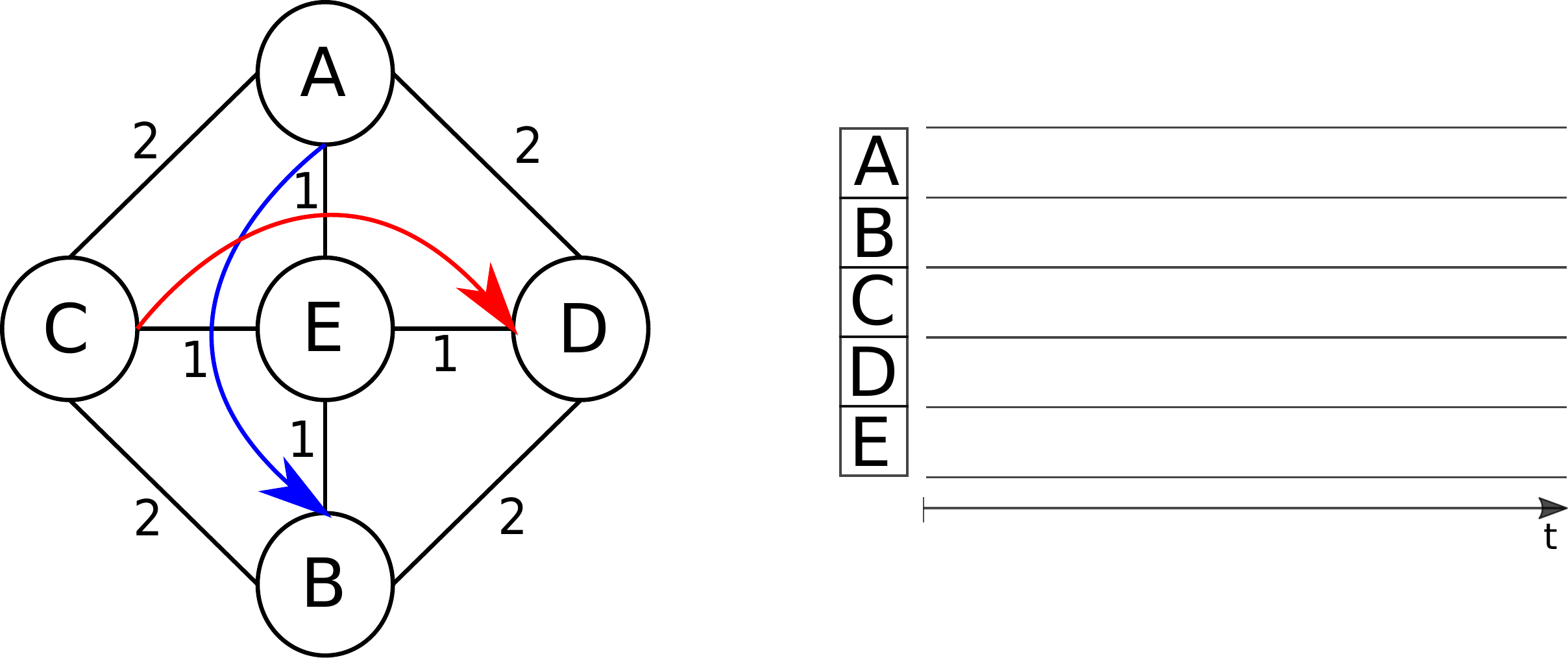}    
        \label{fig:mors_base}
    \end{subfigure}
    \begin{subfigure}[t]{0.50\linewidth}
   	\subcaption{CARP example: path of first robot}
        \includegraphics[width=\linewidth]{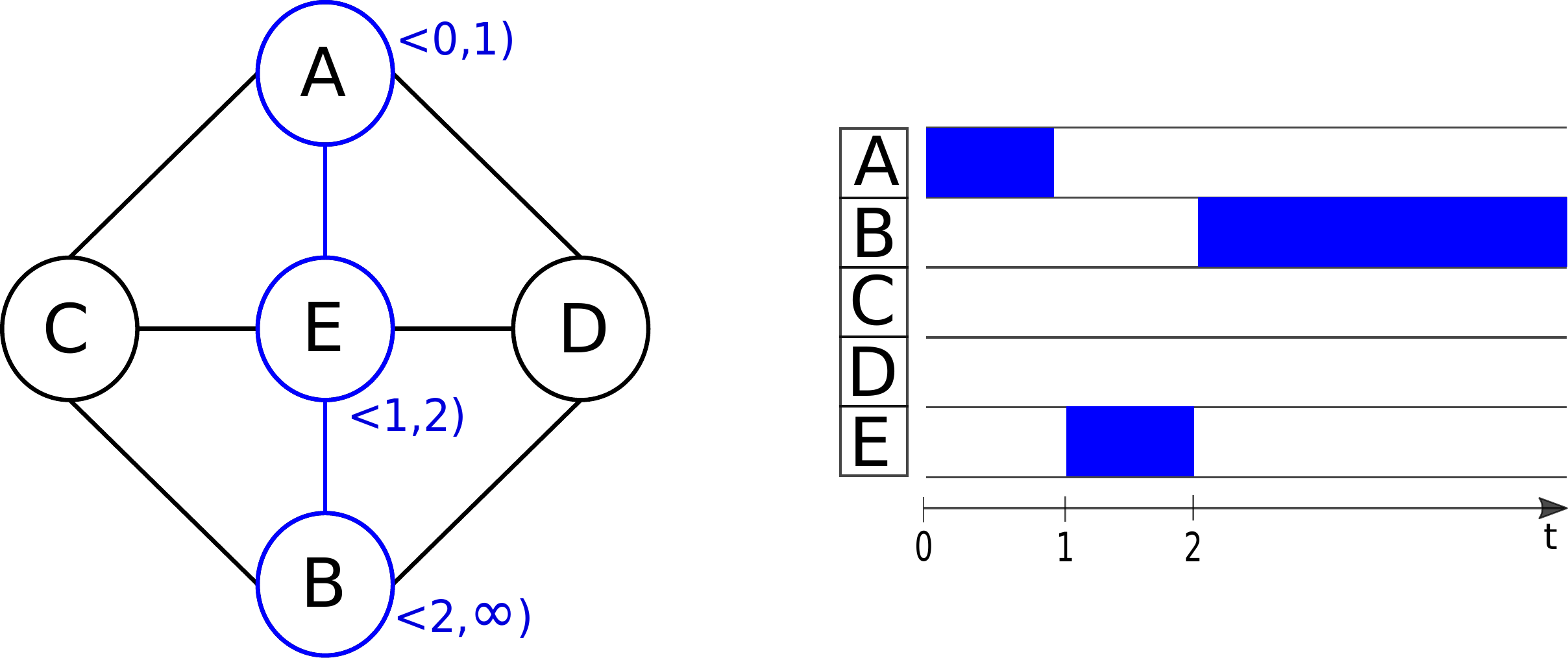}    
        \label{fig:mors_1}
    \end{subfigure}
    
    \center
    \begin{subfigure}[t]{0.50\linewidth}
   	\subcaption{CARP example: path of second robot}
        \includegraphics[width=\linewidth]{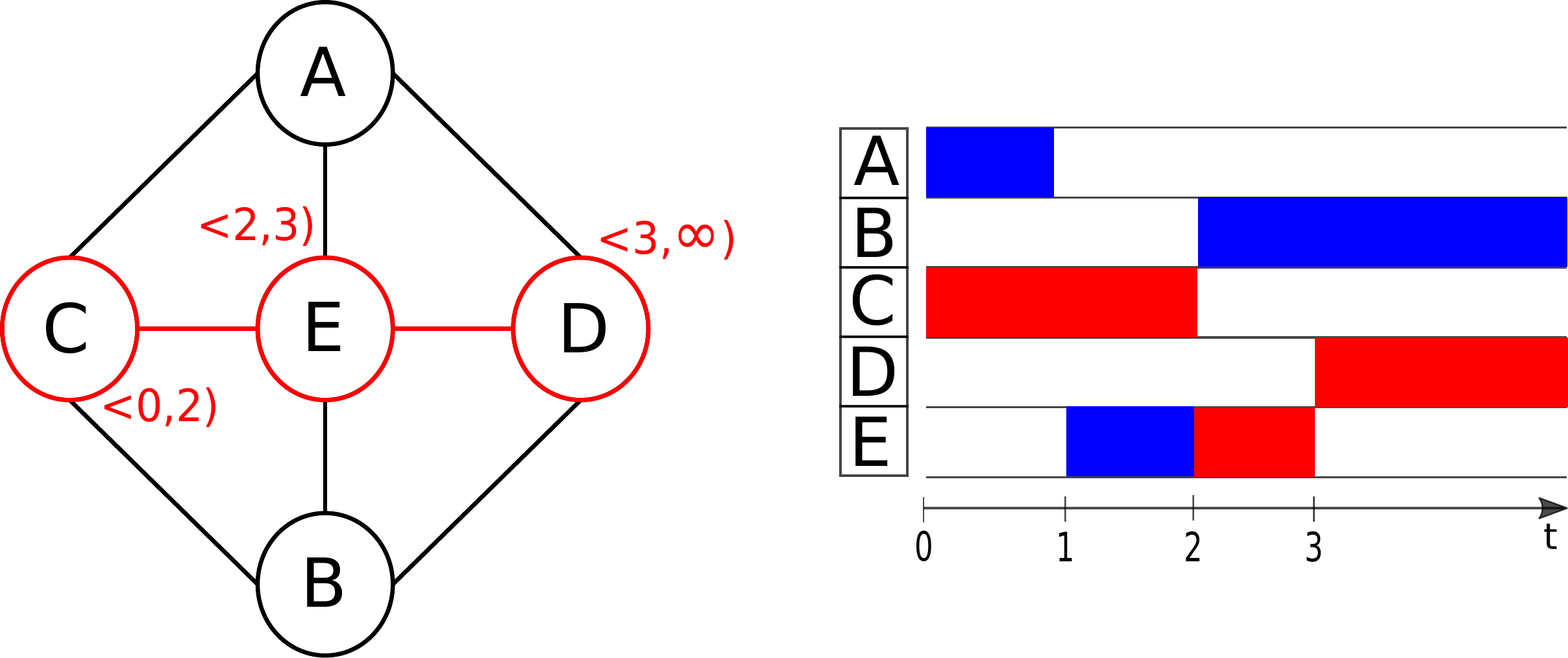}    
        \label{fig:mors_2}
    \end{subfigure}
\caption{Example problem of CARP algorithm\cite{Hvezda2018ICINCO}}
\end{figure*}

An example of pathfinding using CARP algorithm can be seen in Fig.~\ref{fig:mors_base}. Trajectories of two agents need to be found in this example problem. One agent wants to go from the node $A$ to the node $B$ while the second one wants to go from the node $C$ to the node $D$. Because the time window graph is empty at the start and all windows are thus free, the first agent finds its path as a straight route towards its goal. This can be seen in Fig.~\ref{fig:mors_1} along with occupied time windows on the resource graph that were reserved by the first agent. When the second agent plans its path, it is apparent that it cannot go straight towards its goal without pausing. If it chooses to go through the upper path using the node $A$, it would take four time units to reach the node. It can not use the lower path because the node $B$ is taken from time 2 indefinitely. However if the agent waits for one time unit at its starting node $C$ and then move straight towards its goal, the path is free. This resulting plan, as well as the resulting time occupancy for all nodes, can be seen in Fig.~\ref{fig:mors_2}.

\subsubsection{Proposed new steps of the algorithm}

\begin{algorithm}
\caption{Improved MRdRRT algorithm\cite{Hvezda2018ICINCO}}\label{alg:rrt*_overview}
\begin{algorithmic}[1]
\State {$\mathcal{T}.init\left(s\right)$}
\Loop
    \State {$EXPAND\left(\mathcal{T}\right)$}
    \State {$REWIRE\left(\mathcal{T},v'\right)$}
    \State {$\mathcal{P}\leftarrow CONNECT\_TO\_TARGET\left(\mathcal{T},t\right)$}
    \If {$not\_empty(\mathcal{P})$}
        \Return $RETRIEVE\_PATH\left(\mathcal{T},\mathcal{P}\right)$
    \EndIf
\EndLoop
\end{algorithmic}
\end{algorithm}

RRT* algorithm \cite{rrt*} introduced new steps to the original RRT algorithm that enabled it to be asymptotically optimal. 
Our last modifications to the MRdRRT algorithm take inspiration from RRT* to implement the improved expansion step and rewiring step (see Alg.~\ref{alg:rrt*_overview}) in order to increase the quality of the obtained plans.

The tree is initialized with a single node that represents the starting positions of agents (line 1).
The start of the main loop of the algorithm starts consists of the modified expansion phase (line 3).
Once a new node is added to the tree, the rewiring step is called (line 4) which attempts to revise the structure of the tree to improve the length of the path from the added point to the tree root.
The following step is the call to the local connector that checks whether it is possible to connect the newly added point to the goal configuration.
If the local connector is successful, the algorithm terminates and returns the found path as a result.

The main change to the expansion phase, Alg.~\ref{alg:mrmrrt*_expand}, is that after the random sample $u$ is generated (line 1), a set of $N$ nearest neighbours of $u$ in the tree $\mathcal{T}$ is selected.
A new candidate point $v'$ is then generated from each nearest neighbour (lines 6-11) using oracle $\mathcal{O}_D$, but is not added to the tree.
Each candidate point $v'$ is then checked for the distance travelled from the root of the tree $\mathcal{T}$ and only the node that minimizes this distance is connected to the tree to its corresponding predecessor.

This step is the equivalent of similar step in the RRT* algorithm where the algorithm generates new point and then checks all points in a given radius around this point for the best predecessor, meaning a predecessor that minimizes distance through the tree to the root of the tree.
In the multi-agent discrete scenario, it was necessary to adjust this step (Alg. \ref{alg:mrmrrt*_expand}) because the computational requirements would be much higher as the points in given radius would not be able to be connected directly to the new point, because they might not be direct neighbours in the composite configuration space.
This issue would result in the need to use the local connector on each point in the radius.

\begin{algorithm}
\caption{Improved MRdRRT EXPAND$\left(\mathcal{T},r\right)$\cite{Hvezda2018ICINCO}}\label{alg:mrmrrt*_expand}
\begin{algorithmic}[1]
    \State {$u \leftarrow RANDOM\_SAMPLE\left(\right)$}
    \State {$NNs \leftarrow getNearestNeighbours(u)$}    
    \State {$v'_{pred} = -1$}
    \State {$d_{best} = \infty$}
    \State {$v'_{best} = \emptyset$}
    \For {$c \in NNs$}
        \State {$v' \leftarrow \mathcal{O}_D\left(c,u\right)$}
        \If {$l_{\mathcal{T}}\left(c\right) + \delta\left(c,v'\right) < d_{best}$}
            \State {$d_{best} = l_{\mathcal{T}}\left(c\right) + \delta\left(c,v'\right)$}
            \State {$v'_{pred} = c$}
            \State {$v'_{best} = v'$}
        \EndIf
    \EndFor
    \State {$\mathcal{T}.add\_vertex\left(v'_{best}\right)$}
    \State {$\mathcal{T}.add\_edge\left(v'_{pred},v'_{best}\right)$}
\end{algorithmic}
\end{algorithm}

\begin{figure}[htb]
    \begin{subfigure}[b]{\columnwidth}
            \centering
            \includegraphics[width=0.44\columnwidth]{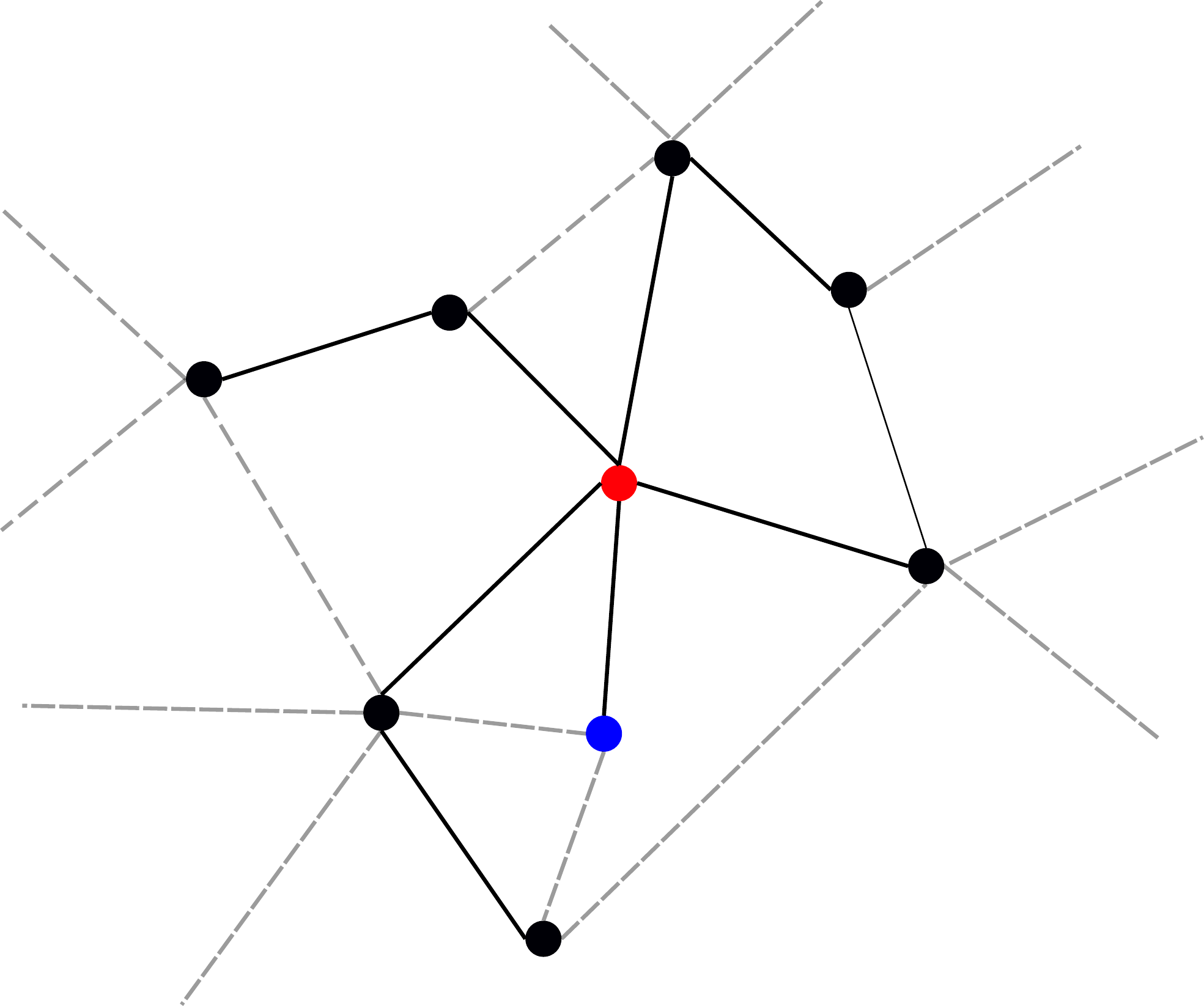}
            \caption{Initial state of the tree. The root node is coloured red, while the newly added node $v'$ is coloured blue.}
            \label{fig:rewire_1}
    \end{subfigure}  

    \bigskip    

    \begin{subfigure}[b]{\columnwidth}
            \centering
            \includegraphics[width=0.44\columnwidth]{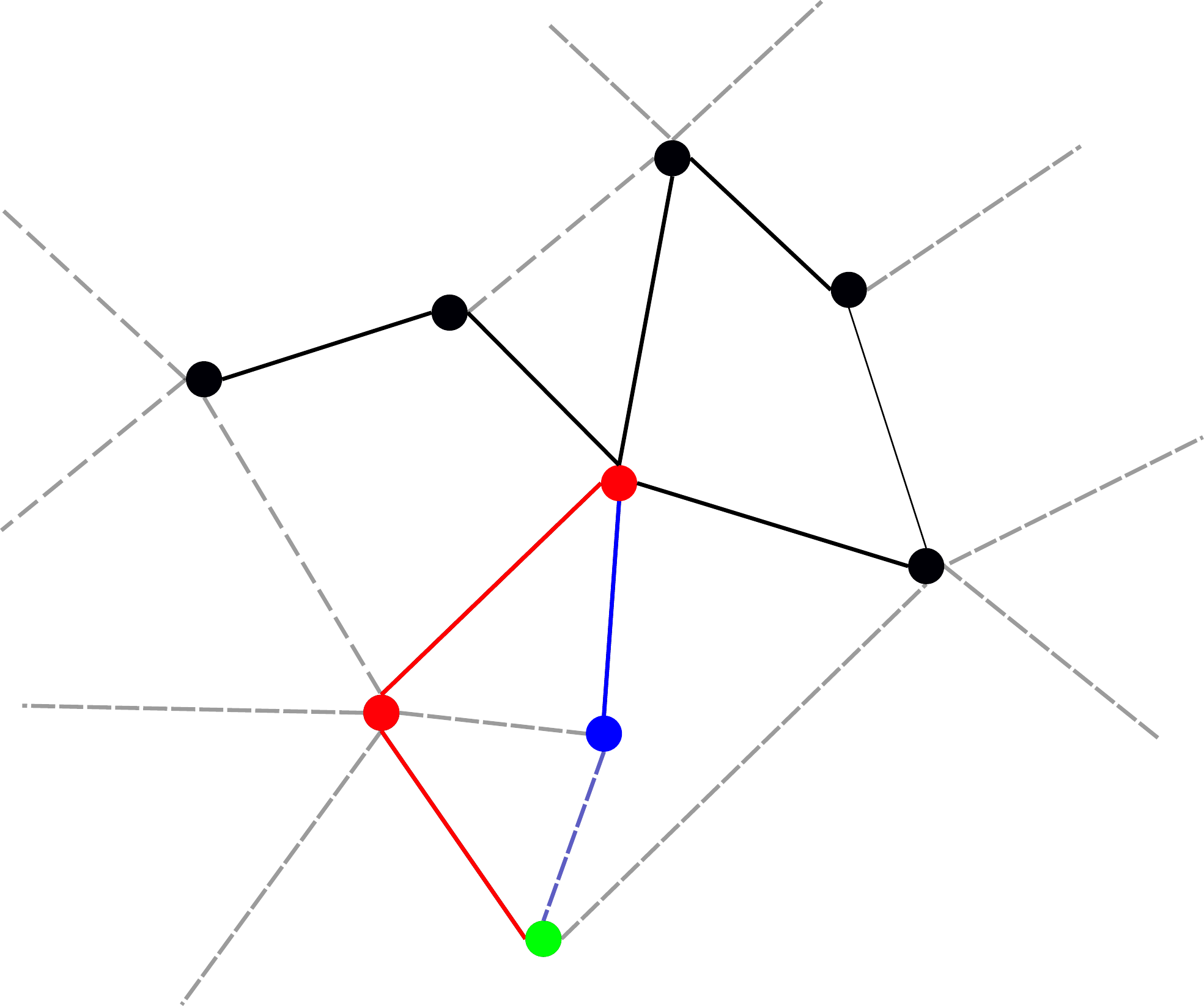}
            \caption{The path through the newly added node $v'$(blue) to one of its nearest neighbors (green) is shorter than the current path to this node (red).}
            \label{fig:rewire_2}
    \end{subfigure}

    \bigskip    
    
    \begin{subfigure}[b]{\columnwidth}
            \centering
            \includegraphics[width=0.44\columnwidth]{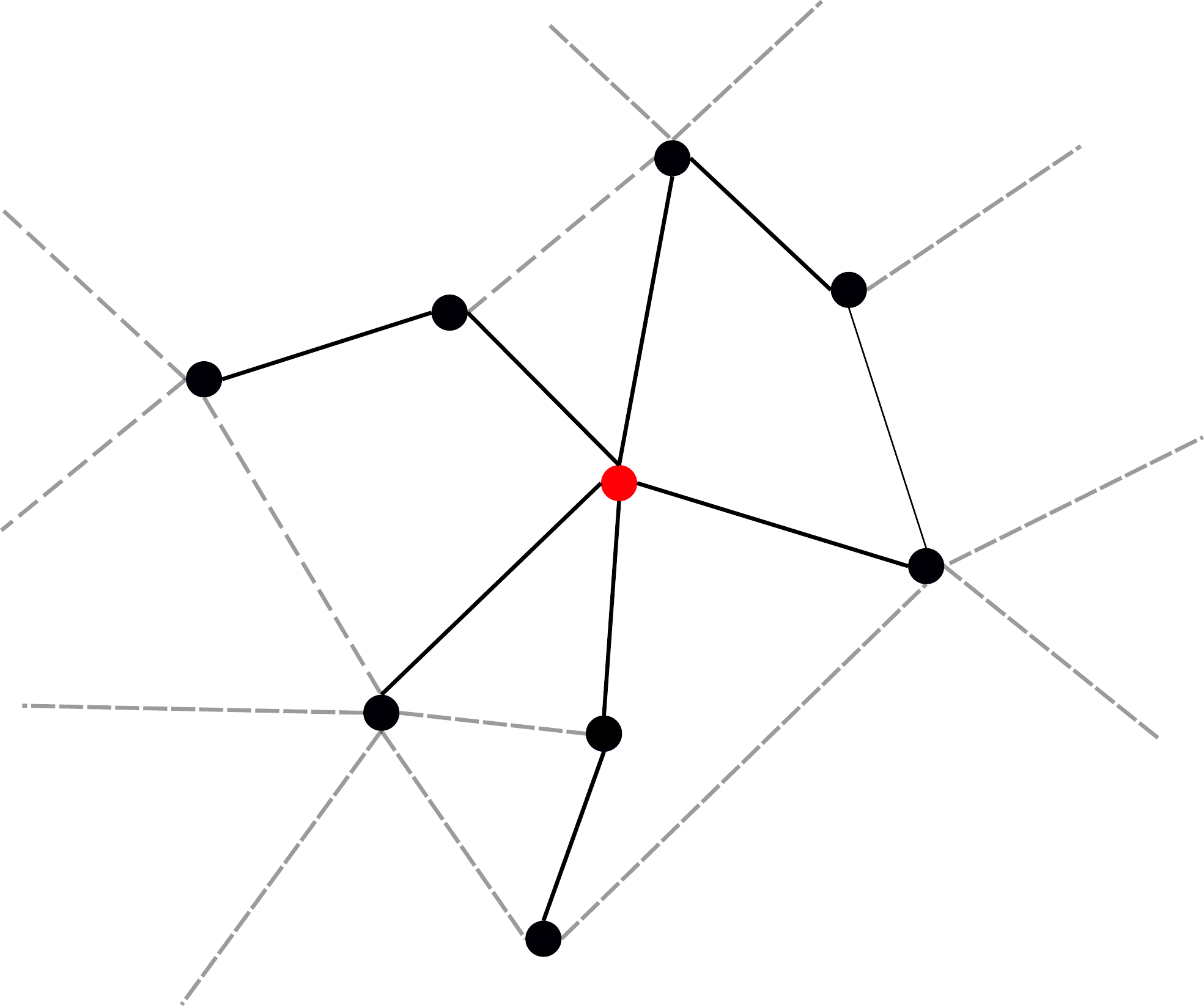}
            \caption{The tree is revised.}
            \label{fig:rewire_3}
    \end{subfigure}
    \caption{Example of the rewiring procedure.} 
    \label{fig:rewire}
\end{figure}

The last step inspired by RRT* that was added to the algorithm is the step called rewiring, which locally revises the structure of $\mathcal{T}$ by checking if nodes that are within certain radius $r$ of newly added point would have the path travelled to them from the root of the tree shorter if they had the newly added point $v'$ as their predecessor.
To avoid the search for all points in a given radius, this step was modified for the multi-agent discrete scenario (Alg.~\ref{alg:mrdrrt*_rewire}) by employing $N$ nearest neighbours search of $v'$ to obtain the neighbouring points $c$. 
Because it is highly unlikely that two neighbouring configurations in the tree are direct neighbours in the composite configuration space, it was necessary to use the local connector to obtain paths that connect configurations $v'$ and $c$ (line 3).
In a case the local connector fails to find connecting paths between the two configurations, the considered point is automatically skipped (lines 4-5).
However, if the local connector succeeds in finding a path $p$ between $v'$ and $c$, it is 
checked whether the length of the path from the root to $v'$ concatenated with the path $p$ and the node $c$ is shorter than the distance travelled through $\mathcal{T}$ from the root to $c$(lines 5-7).
If it is shorter, then all nodes of $p$ are added to $\mathcal{T}$.
The first node of $p$ is connected as the successor of $v'$ and the last node of $p$ is chosen as a new predecessor of $c$.
An example of the rewiring step is displayed in Fig.~\ref{fig:rewire}. 


\begin{algorithm}
\caption{REWIRE$\left(\mathcal{T},v'\right)$\cite{Hvezda2018ICINCO}} \label{alg:mrdrrt*_rewire}
    \begin{algorithmic}[1]
        \State {$NNs \leftarrow getNearestNeighbours(v')$}    
        \For {$c \in NNs$}
            \State {$p \leftarrow LOCAL\_CONNECTOR\left(v',c\right)$}
            \If {$p \leftarrow \emptyset$}
                \State {$Continue$}
            \EndIf
            \State {$n \leftarrow LastNode\left(p\right)$}
            \If {$l_{\mathcal{T}}\left(v'\right)+l\left(p\right)+\delta\left(n,c\right) < l_{\mathcal{T}}\left(c\right)$}            
                \State {$\mathcal{T}.add(p)$}                
                \State {$c.predecessor = n$}
            \EndIf
        \EndFor
    \end{algorithmic}
\end{algorithm}

\section{{Experiments}}
\label{sec:experiments}

\begin{figure*}
\label{fig:first_experiment}	
    \begin{subfigure}[t]{0.44\linewidth}
   	\subcaption{Success rate}
        \includegraphics[width=\linewidth]{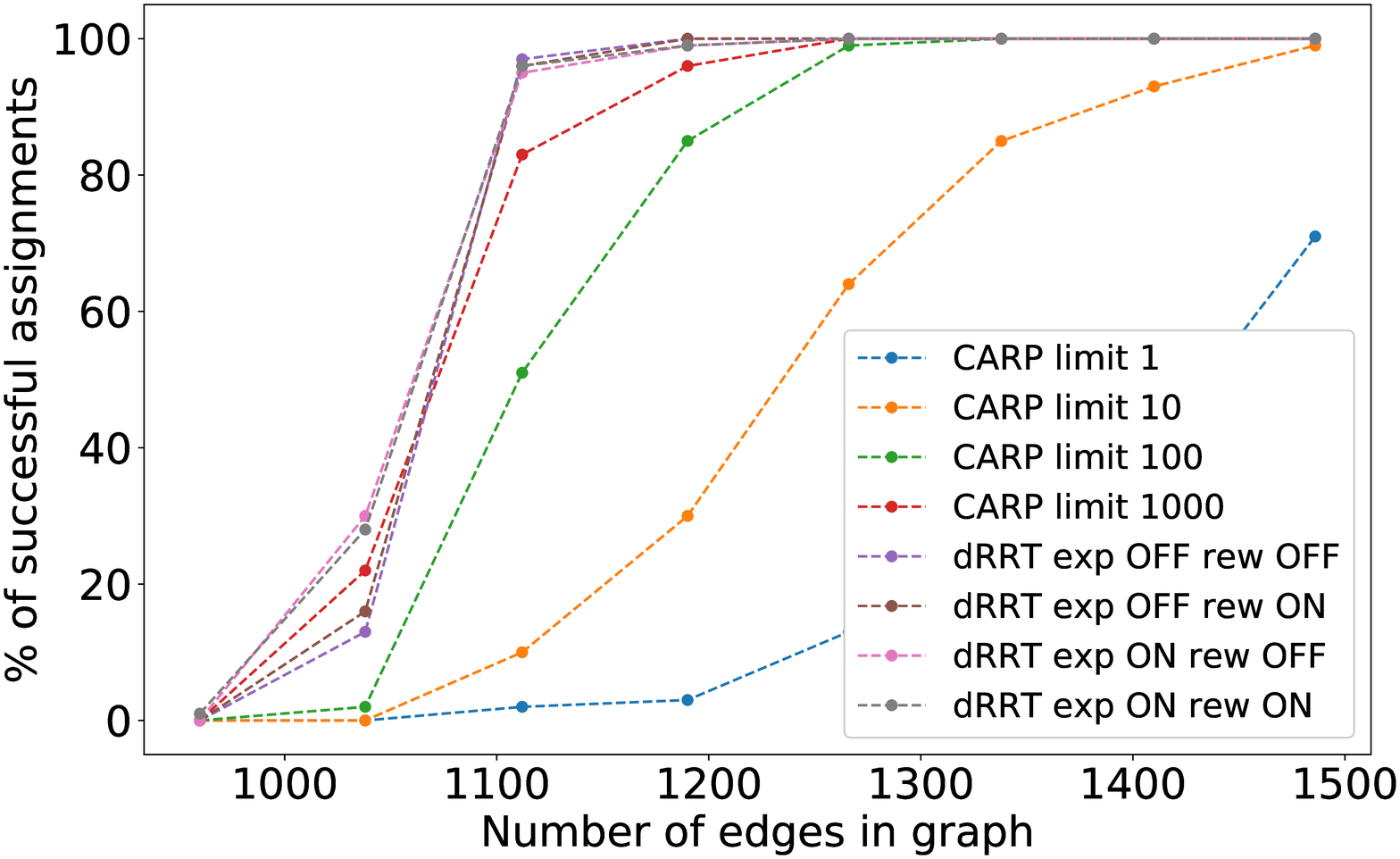}    
        \label{fig:failrate_20}
    \end{subfigure}
    \begin{subfigure}[t]{0.44\linewidth}
    	\subcaption{Median number of plan steps}
        \includegraphics[width=\linewidth]{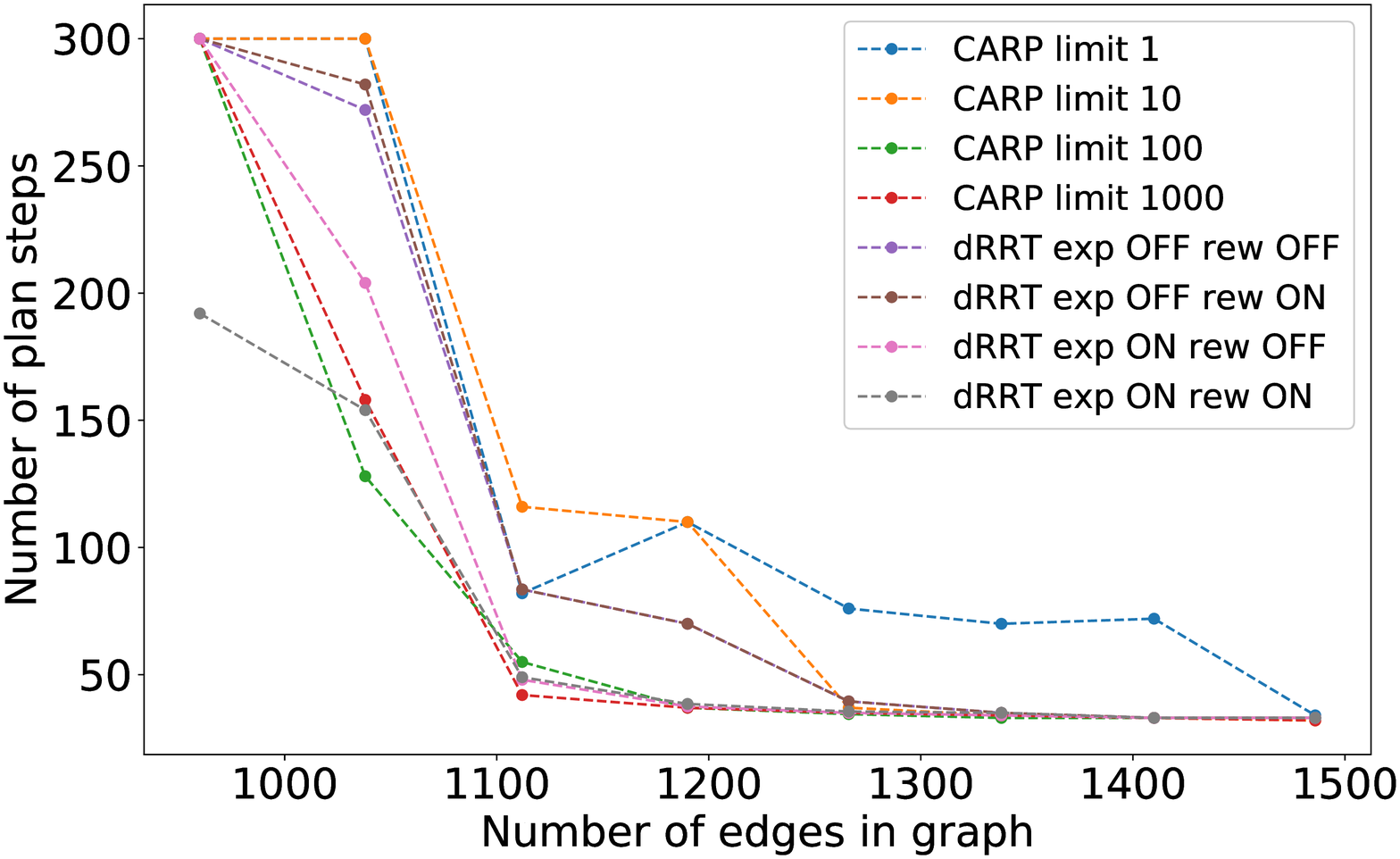}    
        \label{fig:steps_20}
    \end{subfigure}    
    
    \begin{subfigure}[t]{0.44\linewidth}\
		\subcaption{Median iterations}        
        \includegraphics[width=\linewidth]{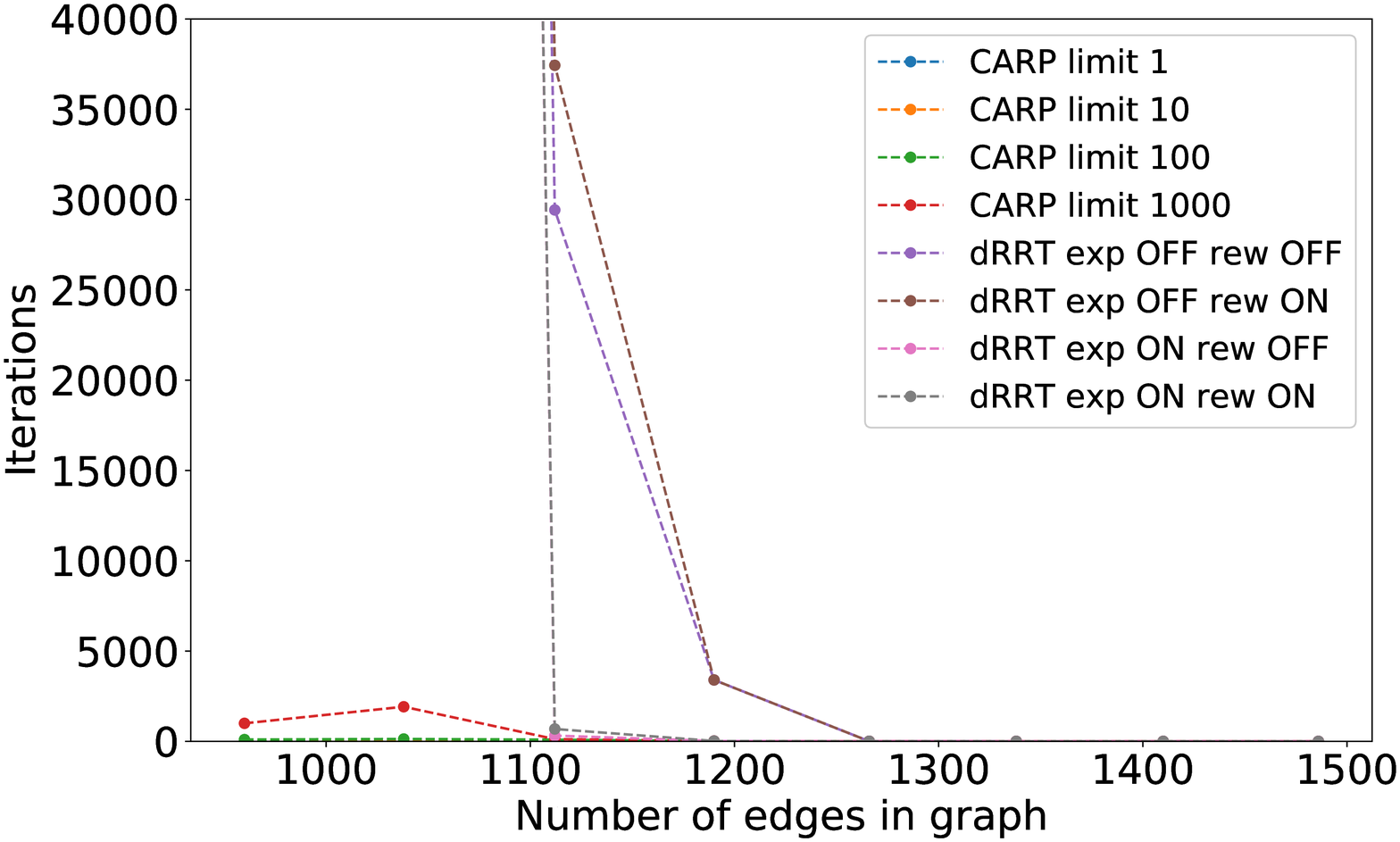}    
        \label{fig:iterations_20}
    \end{subfigure}    
    \begin{subfigure}[t]{0.44\linewidth}
    	\subcaption{Median time to plan}
        \includegraphics[width=\linewidth]{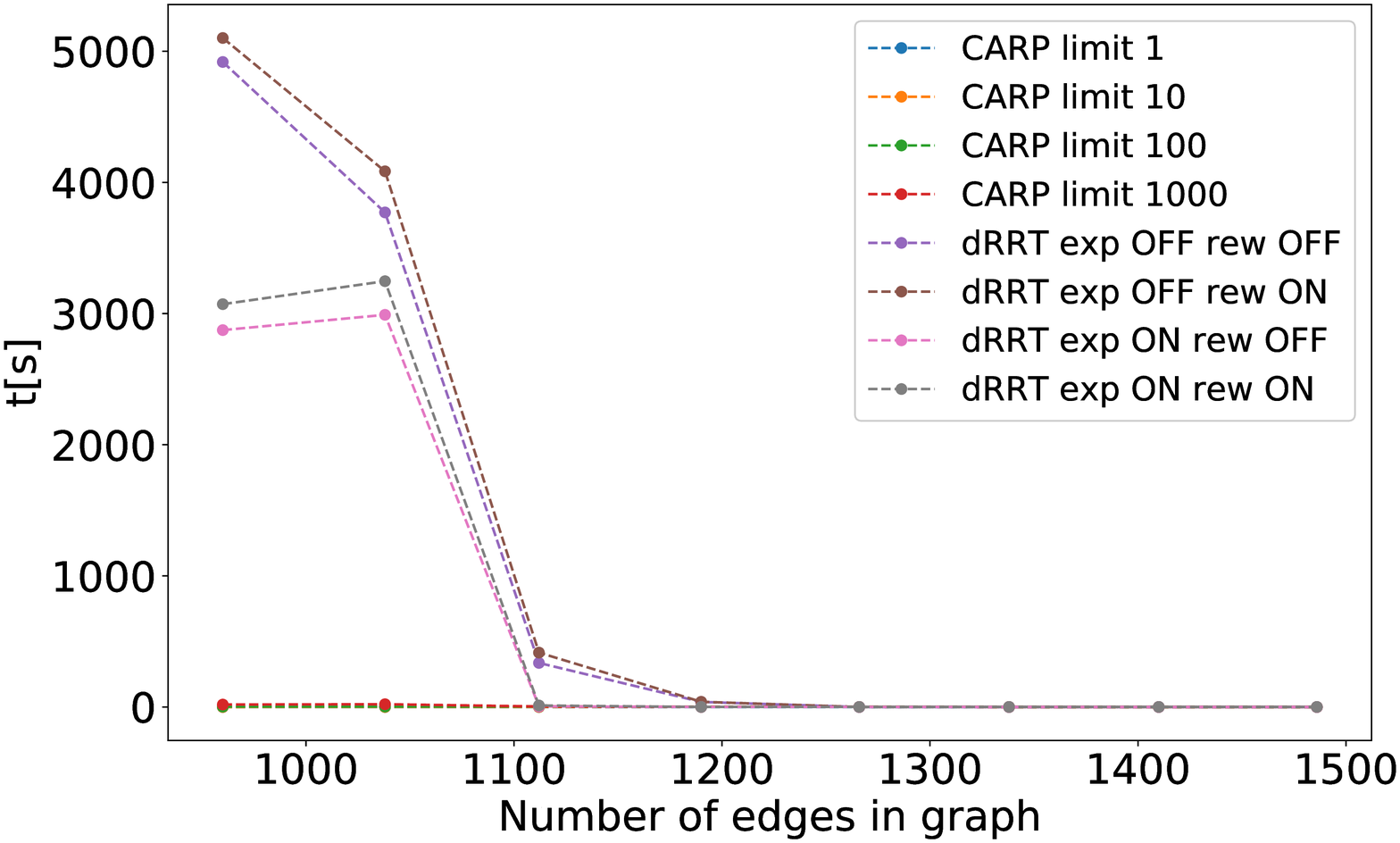}    
        \label{fig:runtime_20}
    \end{subfigure}
    \caption{Results of the first experiment for the maps from $20\times20$ grid}
    \label{fig:first_experiment_20}
\end{figure*}

\begin{figure*}	
    \begin{subfigure}[t]{0.44\linewidth}
   	\subcaption{Success rate}
        \includegraphics[width=\linewidth]{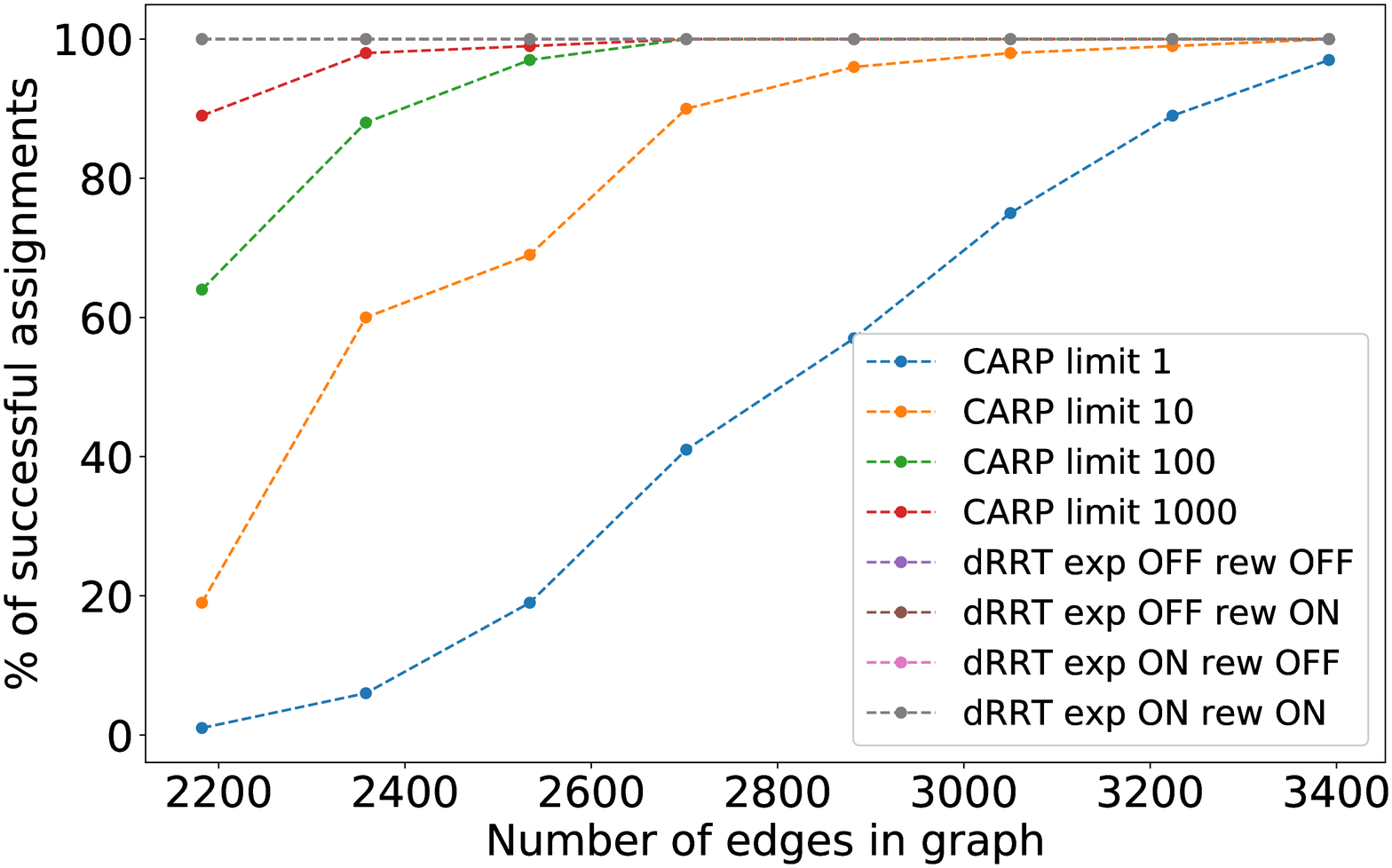}    
        \label{fig:failrate_30}
    \end{subfigure}
    \begin{subfigure}[t]{0.44\linewidth}
    	\subcaption{Median number of plan steps}
        \includegraphics[width=\linewidth]{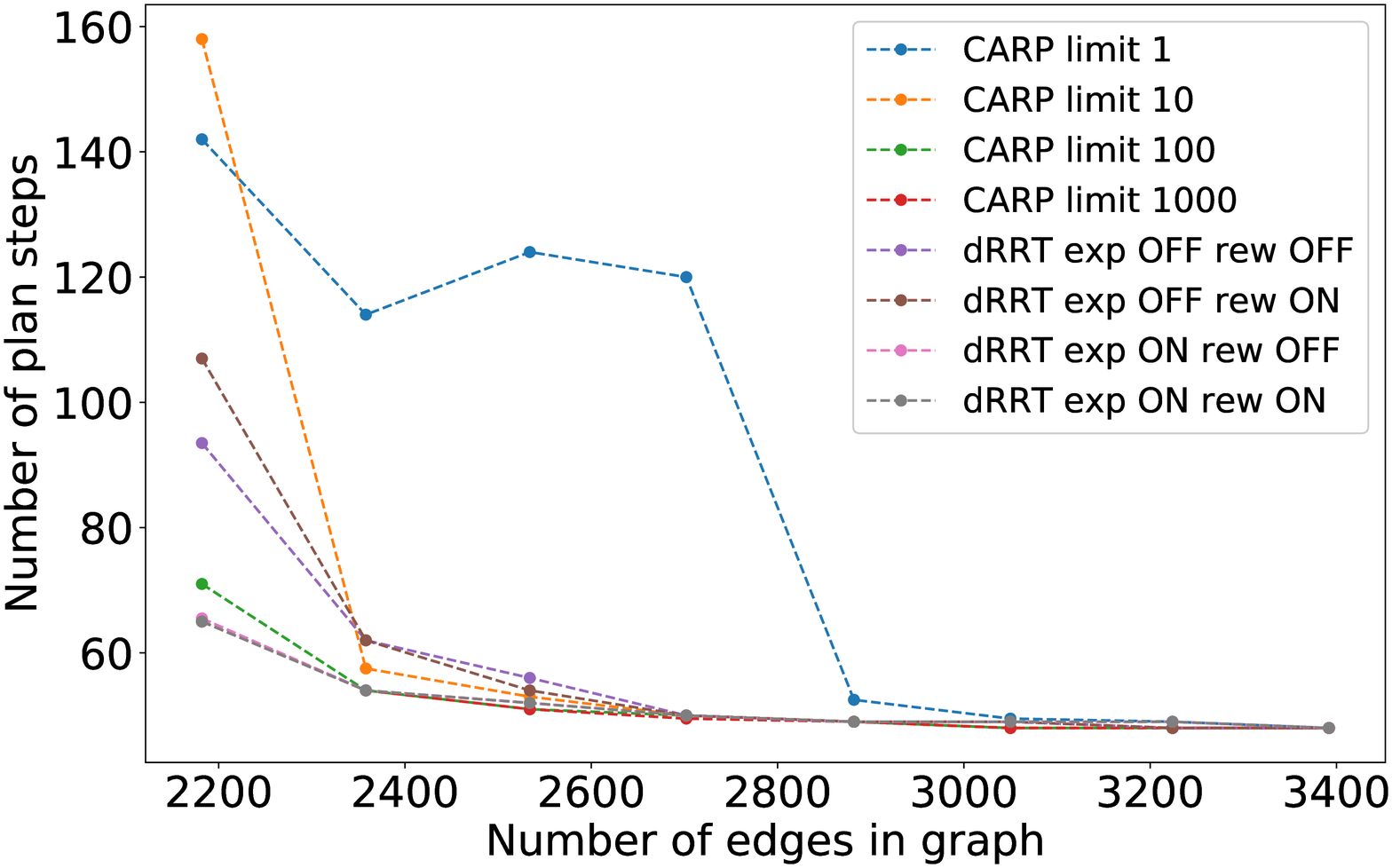}    
        \label{fig:steps_30}
    \end{subfigure}    
    
    \begin{subfigure}[t]{0.44\linewidth}\
		\subcaption{Median iterations}        
        \includegraphics[width=\linewidth]{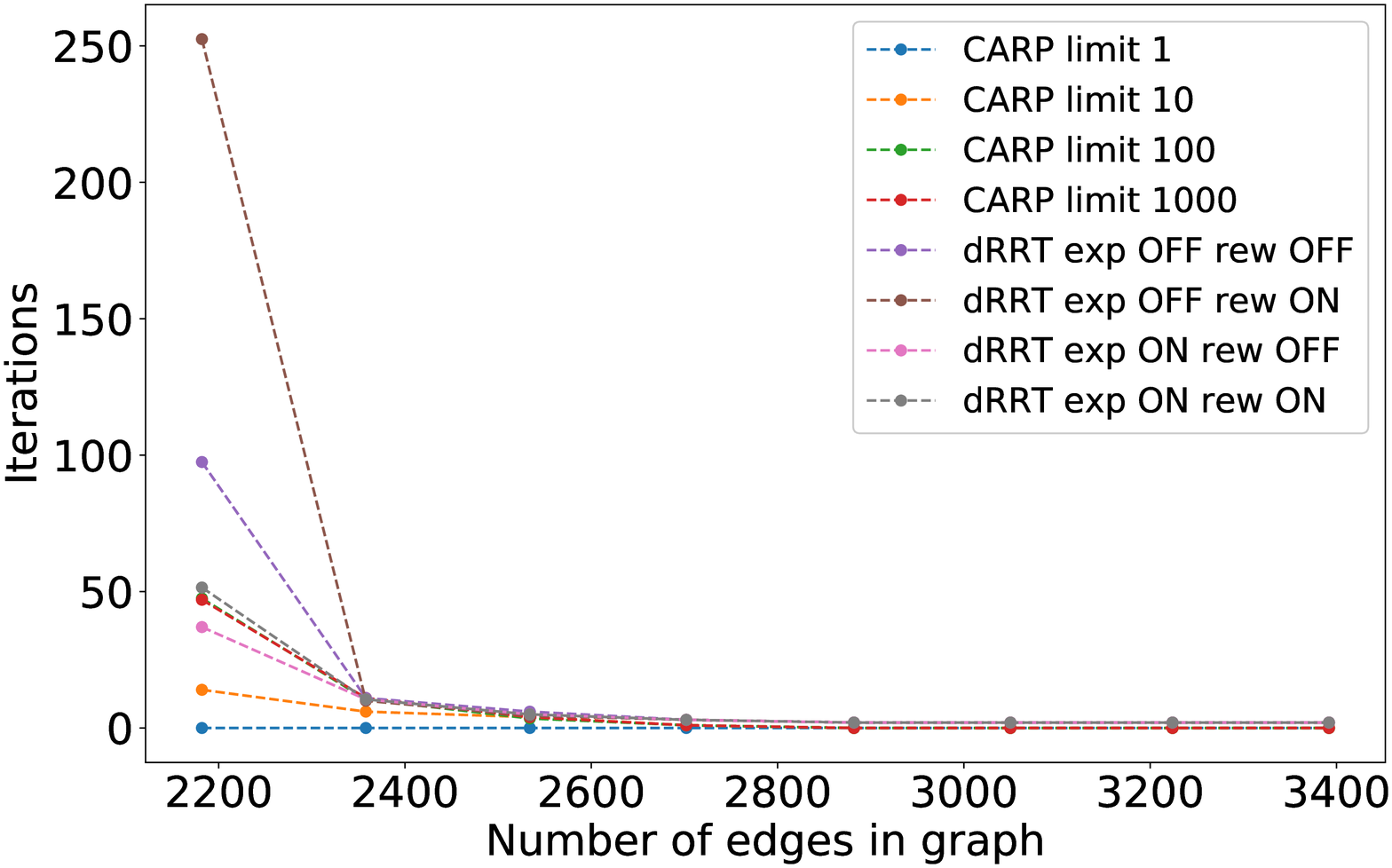}    
        \label{fig:iterations_30}
    \end{subfigure}    
    \begin{subfigure}[t]{0.44\linewidth}
    	\subcaption{Median time to plan}
        \includegraphics[width=\linewidth]{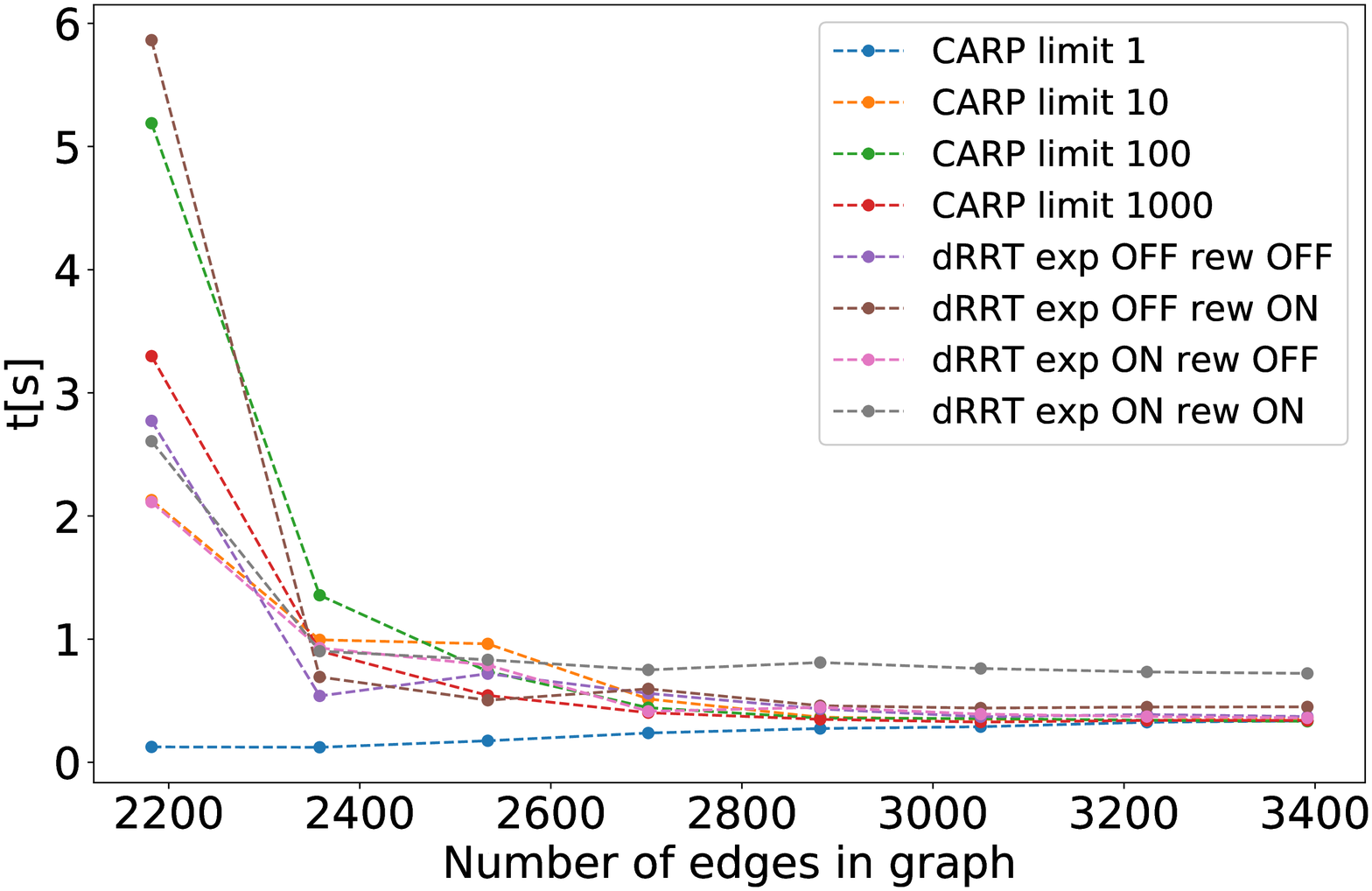}    
        \label{fig:runtime_30}
    \end{subfigure}
    \caption{Results of the first experiment for the maps from $30\times30$ grid}
    \label{fig:first_experiment_30}
\end{figure*}

\begin{figure*}
    \begin{subfigure}[t]{0.44\linewidth}
   	\subcaption{Success rate}
        \includegraphics[width=\linewidth]{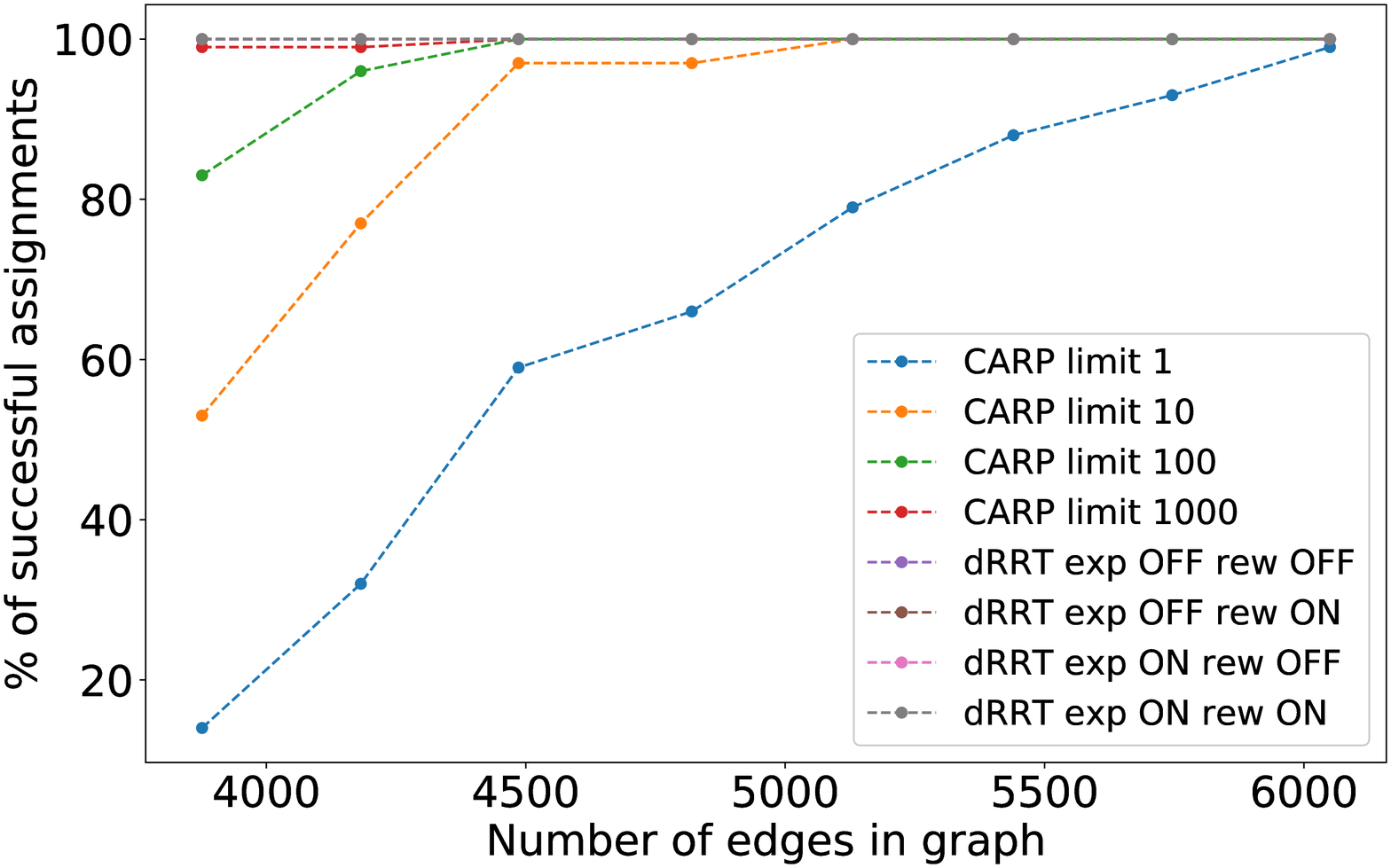}    
        \label{fig:failrate_40}
    \end{subfigure}
    \begin{subfigure}[t]{0.44\linewidth}
    	\subcaption{Median number of plan steps}
        \includegraphics[width=\linewidth]{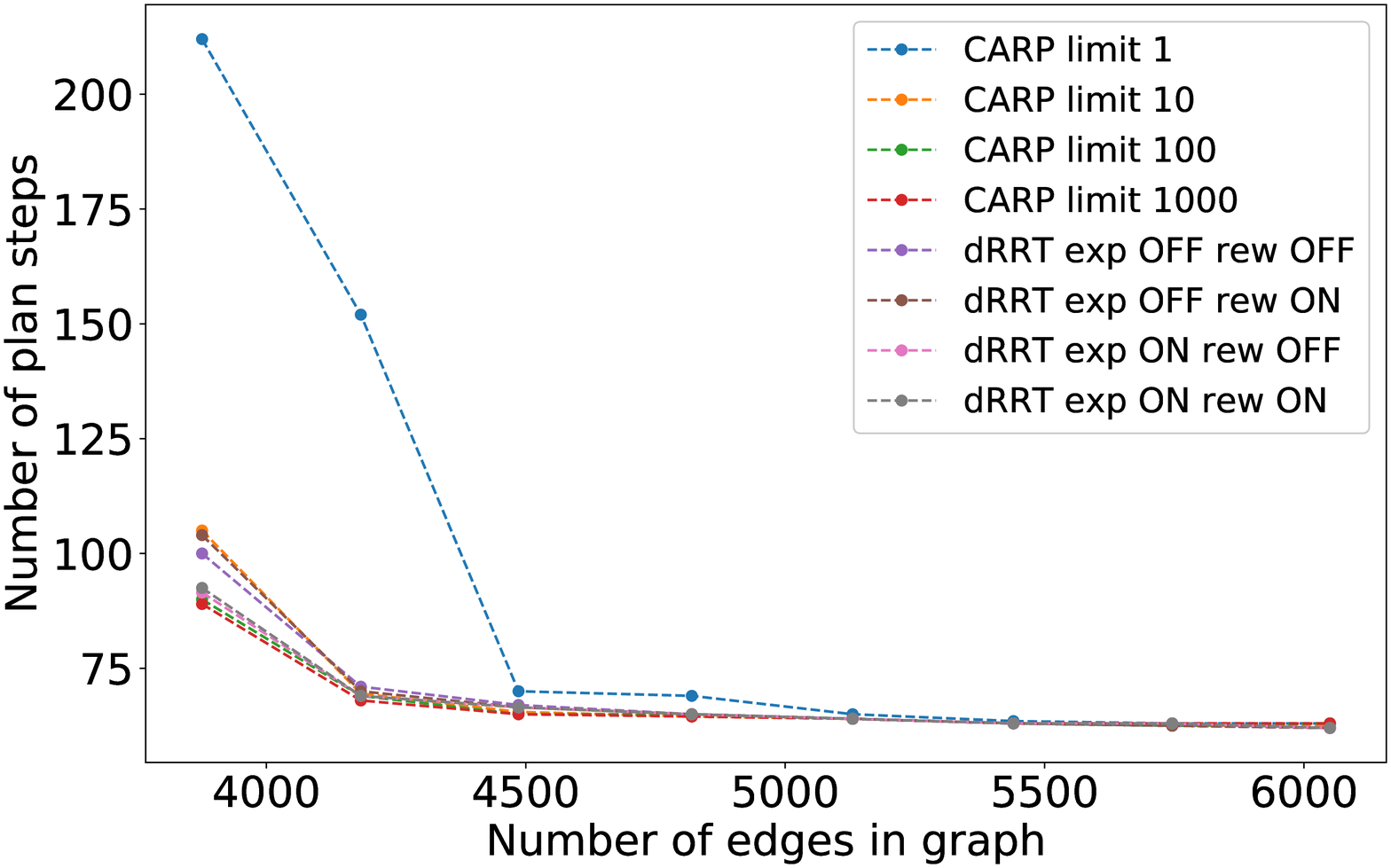}    
        \label{fig:steps_40}
    \end{subfigure}    
    
    \begin{subfigure}[t]{0.44\linewidth}\
		\subcaption{Median iterations}        
        \includegraphics[width=\linewidth]{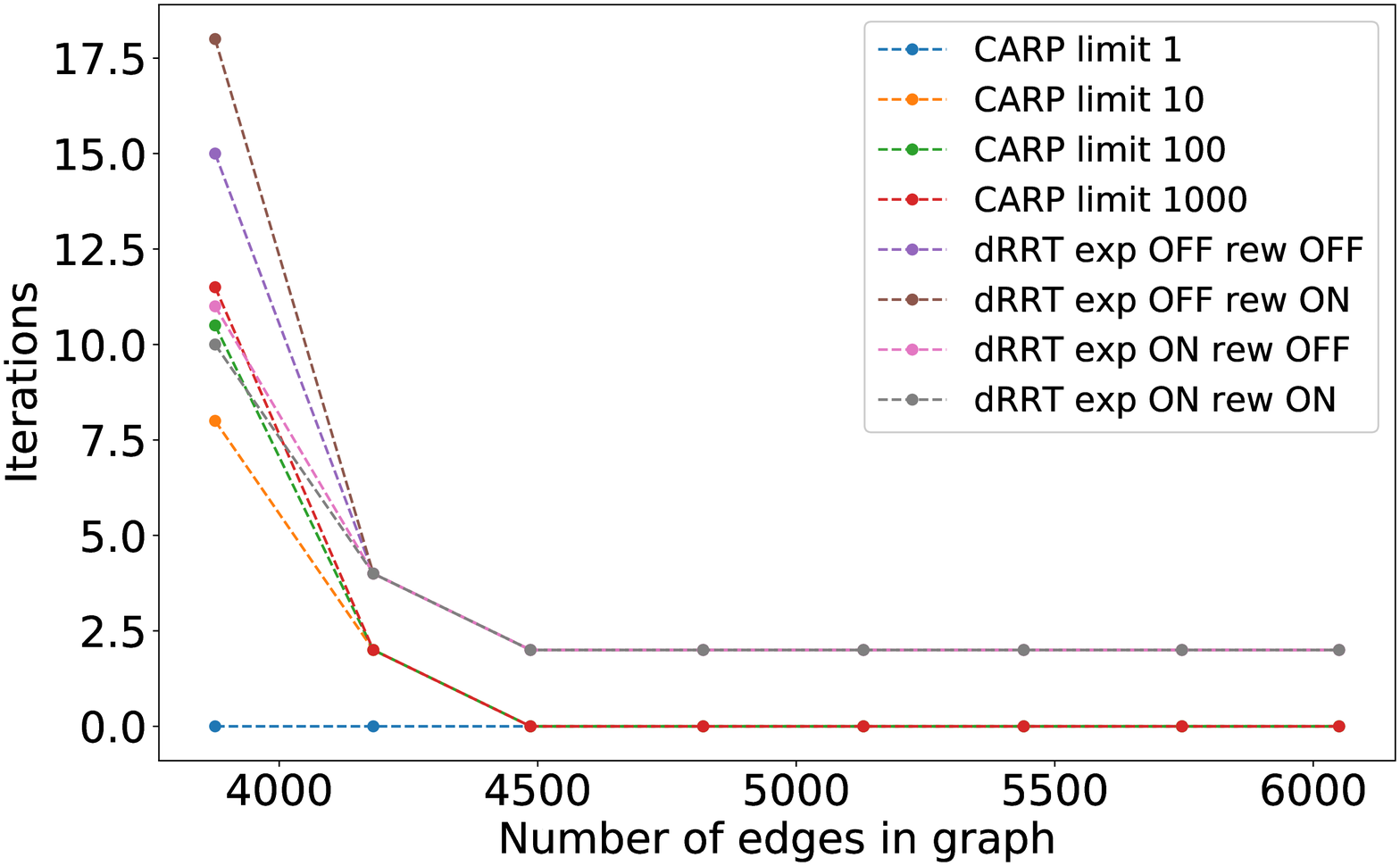}    
        \label{fig:iterations_40}
    \end{subfigure}    
    \begin{subfigure}[t]{0.44\linewidth}
    	\subcaption{Median time to plan}
        \includegraphics[width=\linewidth]{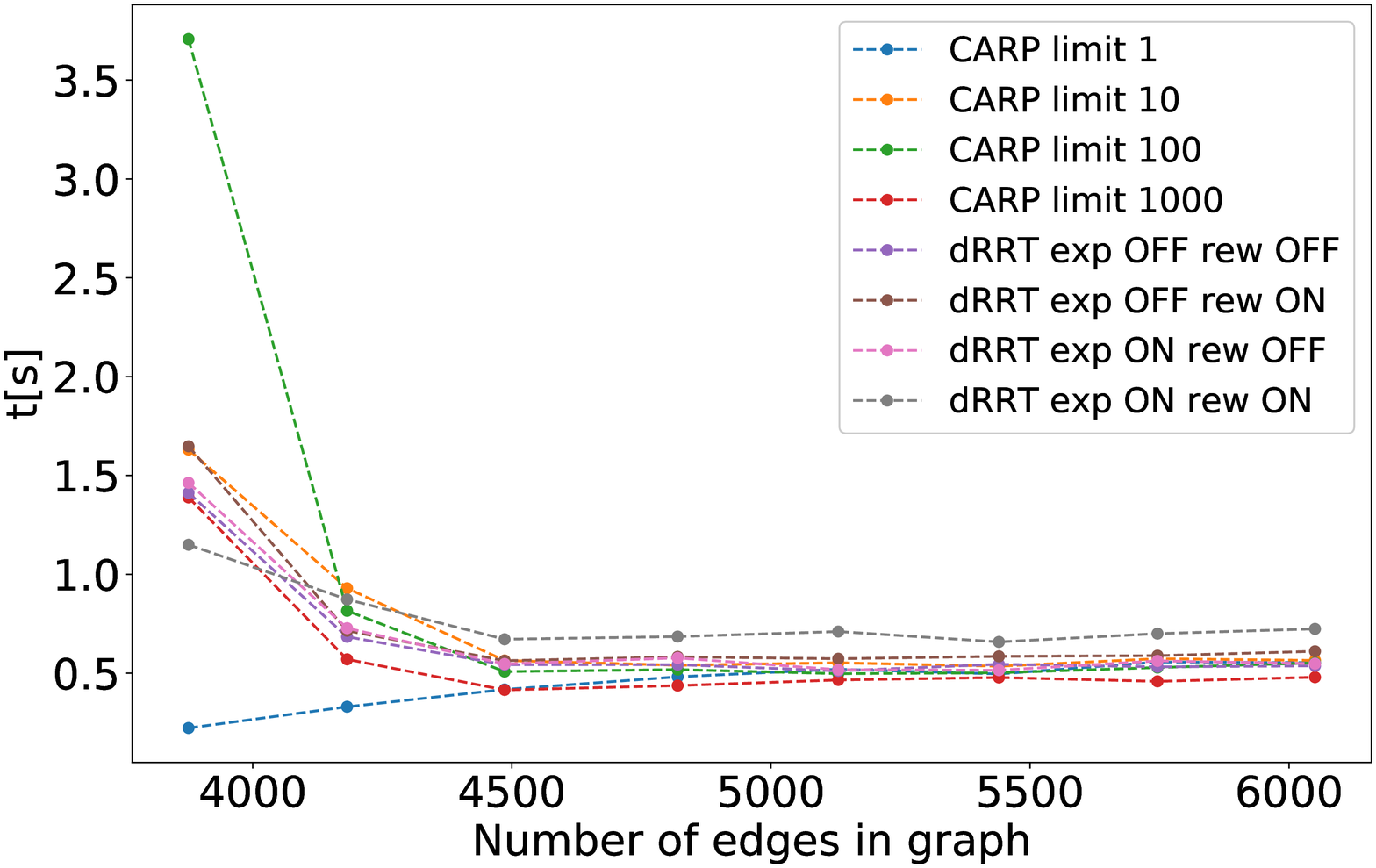}    
        \label{fig:runtime_40}
    \end{subfigure}
    \caption{Results of the first experiment for the maps from $40\times40$ grid}
    \label{fig:first_experiment_40}
\end{figure*}

To show the performance of the proposed algorithm we chose to compare it with the CARP algorithm~\cite{terMors2010}. Because the CARP algorithm is heavily dependent on the ordering in which the agents are planned we have used several variants that differ in the maximal number of allowed shuffles of the ordering. Once CARP finds solution with a given ordering, we consider it succeeded and returned its result. We used 4 such variants with 1, 10, 100 and 1000 maximum possible shuffles of the ordering.

The comparison was done in two separate experiments, each of which contained its own set of maps and assignments. The primary goal was to evaluate the reliability of the algorithm as well as its runtime. To achieve this, we have recorded metrics based on which we have made the comparison of the tested algorithms. These metrics are the following: success rate, length of the resulting plan, sum of individual agent plans in the given plan, runtime of the algorithm and the number of iterations the algorithm needed to find a solution.

\begin{figure*}
    \begin{subfigure}[t]{0.44\linewidth}
        \includegraphics[width=\linewidth]{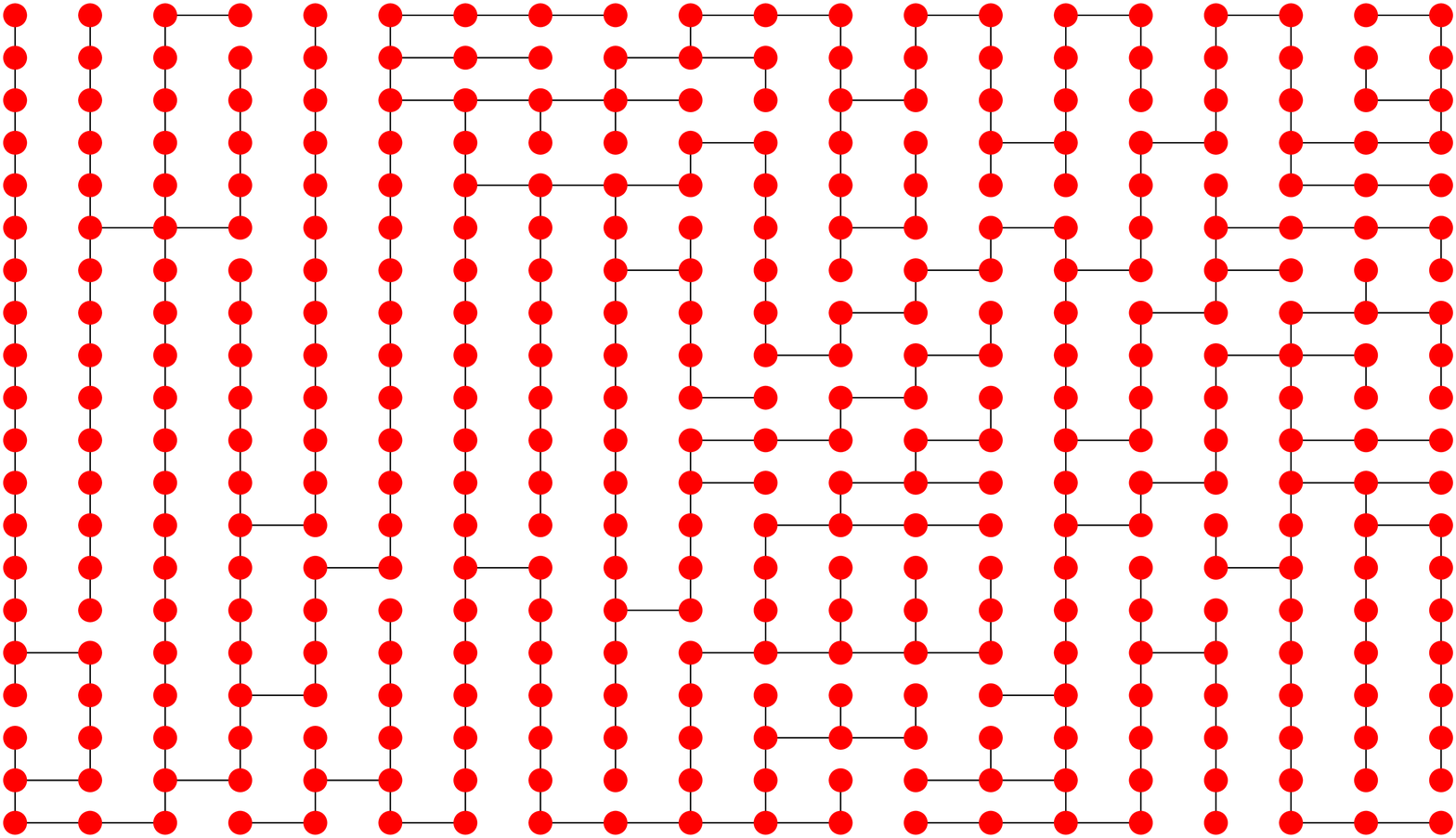}    
        \label{fig:firstExperimentMap_1}
    \end{subfigure}
    \begin{subfigure}[t]{0.44\linewidth}
        \includegraphics[width=\linewidth]{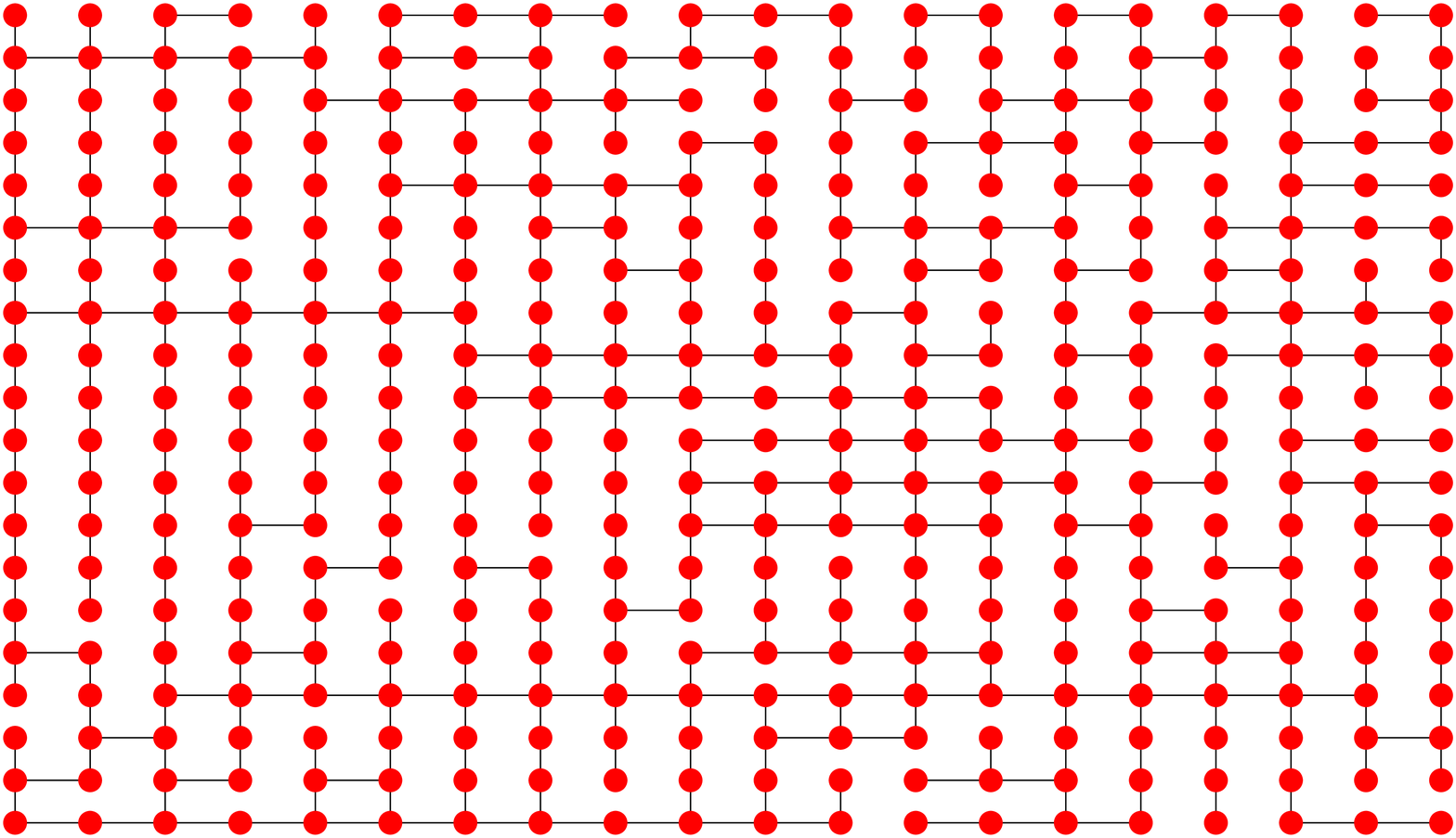}    
        \label{fig:firstExperimentMap_2}
    \end{subfigure}    
    
    \center
    \begin{subfigure}[t]{0.44\linewidth}\
        \includegraphics[width=\linewidth]{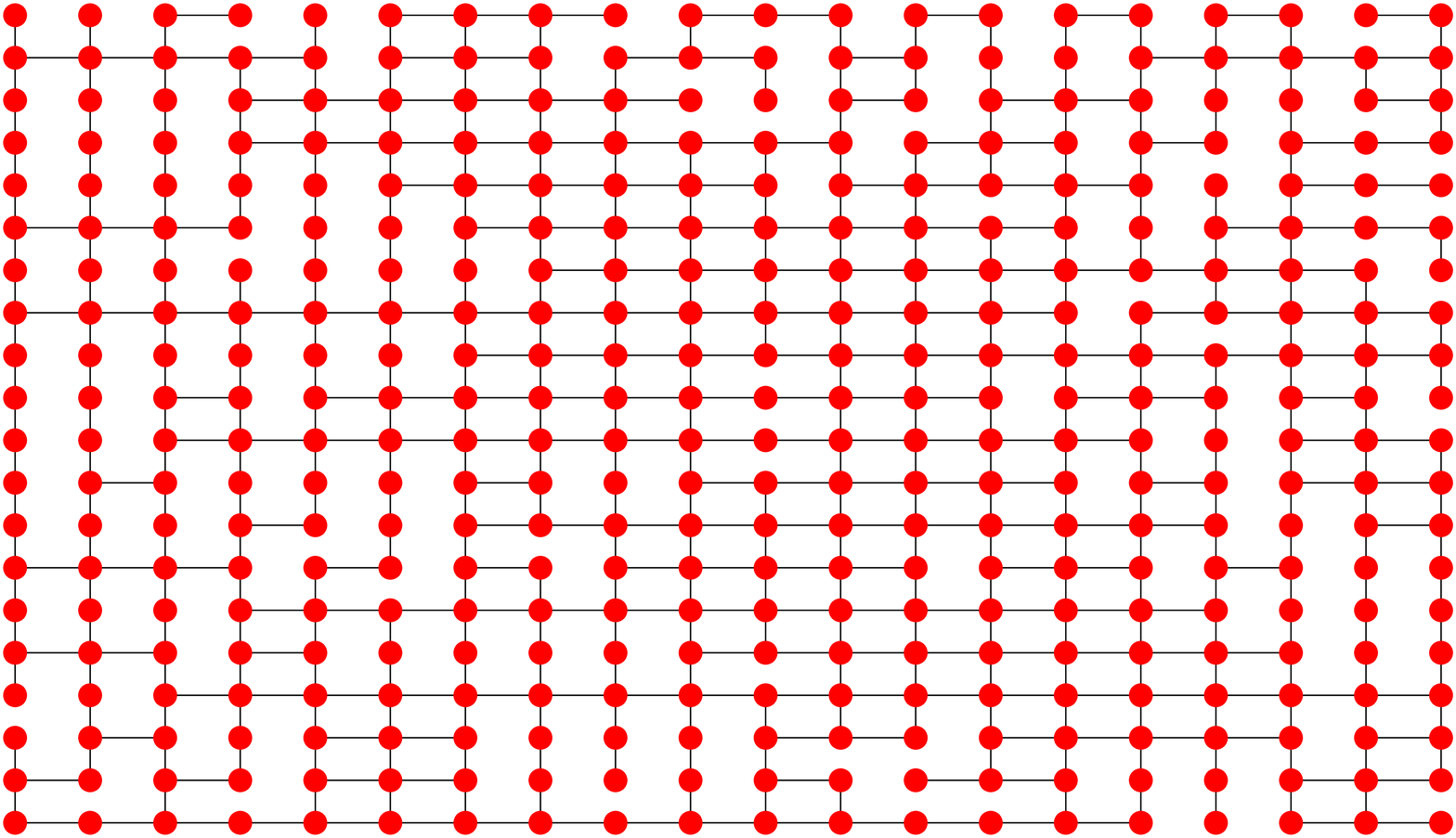}    
        \label{fig:firstExperimentMap_3}
    \end{subfigure}    

    \caption{Example of maps from the first experiment}
    \label{fig:first_experiment_maps}	
\end{figure*}

The first experiment was performed with the aim to evaluate how the algorithm performs on maps with various densities, e.g.  with a sequence of maps with an increasing number of edges given the same set of nodes.
The set was created by initially generating grid maps with preset a number of nodes. 
More precisely, we have generated three such grids with $20\times20$, $30\times30$ and $40\times40$ nodes. In the next step, a minimum spanning tree for each grid was determined, which created the base map. 
Once the minimum spanning trees have been created, 9 more maps have been created for each spanning tree by iteratively adding a fixed number of edges from the original grid. This procedure resulted in 30 different maps, 10 for each grid. An example of these maps can be seen in Fig.~\ref{fig:first_experiment_maps}.

Furthermore, for each of the grids 100 different assignments were generated assuming a fleet of 100 robots. These assignments were created by randomly sampling pairs of nodes from the the given grid.

In addition, our algorithm was tested in 4 different variants that differ by either enabling or disabling two main new components -- improved expansion and rewiring.

As the algorithms could fail, the results from failed planning attempts were handled such that if any of the 100 assignments was successful, all failed attempts were set as 2 times the worst result. In case none of the 100 assignments was planned successfully, all failed attempts were set to arbitrary high value, in our case 100000. The graphs then show median value over the 100 assignments for each map of each grid. In the case of runtime measurements, the time was measured from the start of the algorithm until result was returned, even if the result was a failure. Moreover, the proposed algorithm was given 500000 maximum iterations to find a plan, otherwise the attempt was considered failed in which case the 500000 was reported as the number of iterations. For the CARP algorithm, if a failure to plan was detected, the number of iterations was set to the given maximum limit of shuffles. For example, if CARP with the limit of 1000 shuffles failed, it reported the number of iterations was 1000.

The results of the experiment can be seen in Figs.~\ref{fig:first_experiment_20}, \ref{fig:first_experiment_30}, and \ref{fig:first_experiment_40}. 
Note that results for the first two maps are not shown in the graphs, because none of the algorithms was able to find a solution for any assignment.
The first thing to notice is that the proposed approach shows much higher success rate in all its variants than all CARP variants in Figs.~\ref{fig:failrate_20}, \ref{fig:failrate_30} and \ref{fig:failrate_40}. Note that the two versions with the improved expansion turned on provided a higher success rate on the map created from $20\times20$ grid with less edges than the two versions with this component turned off.
Upon inspecting the length on the plan in the Figs.~\ref{fig:steps_20}, \ref{fig:steps_30}, and \ref{fig:steps_40} it can be seen that our approach provides comparable results with CARP variants in all versions with the versions where the improved expansion was turned on providing better results on more sparse maps while also requiring less computational time to obtain the result. 
The worst results are provided on the more sparse maps created from the $20\times20$ grid as can be seen for example in Fig.~\ref{fig:iterations_20} where the variants of the proposed algorithm were not able to find a solution in more than half the assignments for the maps with 960 and 1038 edges, and thus the median number of iterations was set to the preset limit of 500000.

\begin{figure}[htb]
    \centering
    \begin{subfigure}[t]{0.4\columnwidth}
            \centering
            \includegraphics[width=\columnwidth]{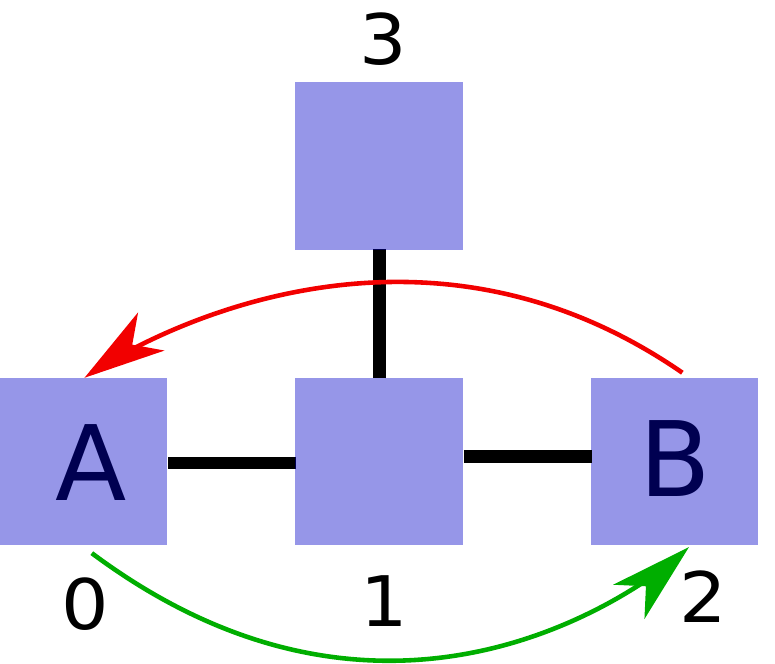}            
            \caption{Base problem.}
            \label{fig:mapgen_base}
    \end{subfigure} 
    \hfil
    \begin{subfigure}[t]{0.4\columnwidth}
            \centering
            \includegraphics[width=\columnwidth]{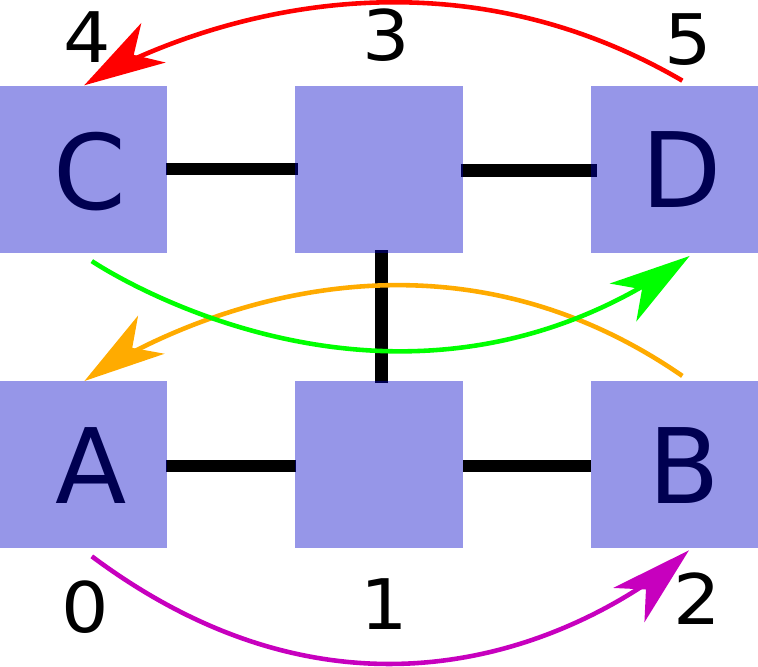}
            \caption{First type of map expansion}
            \label{fig:mapgen_1}
    \end{subfigure}
    
    \bigskip\bigskip

    \begin{subfigure}[b]{\columnwidth}
            \centering
            \includegraphics[width=0.76\columnwidth]{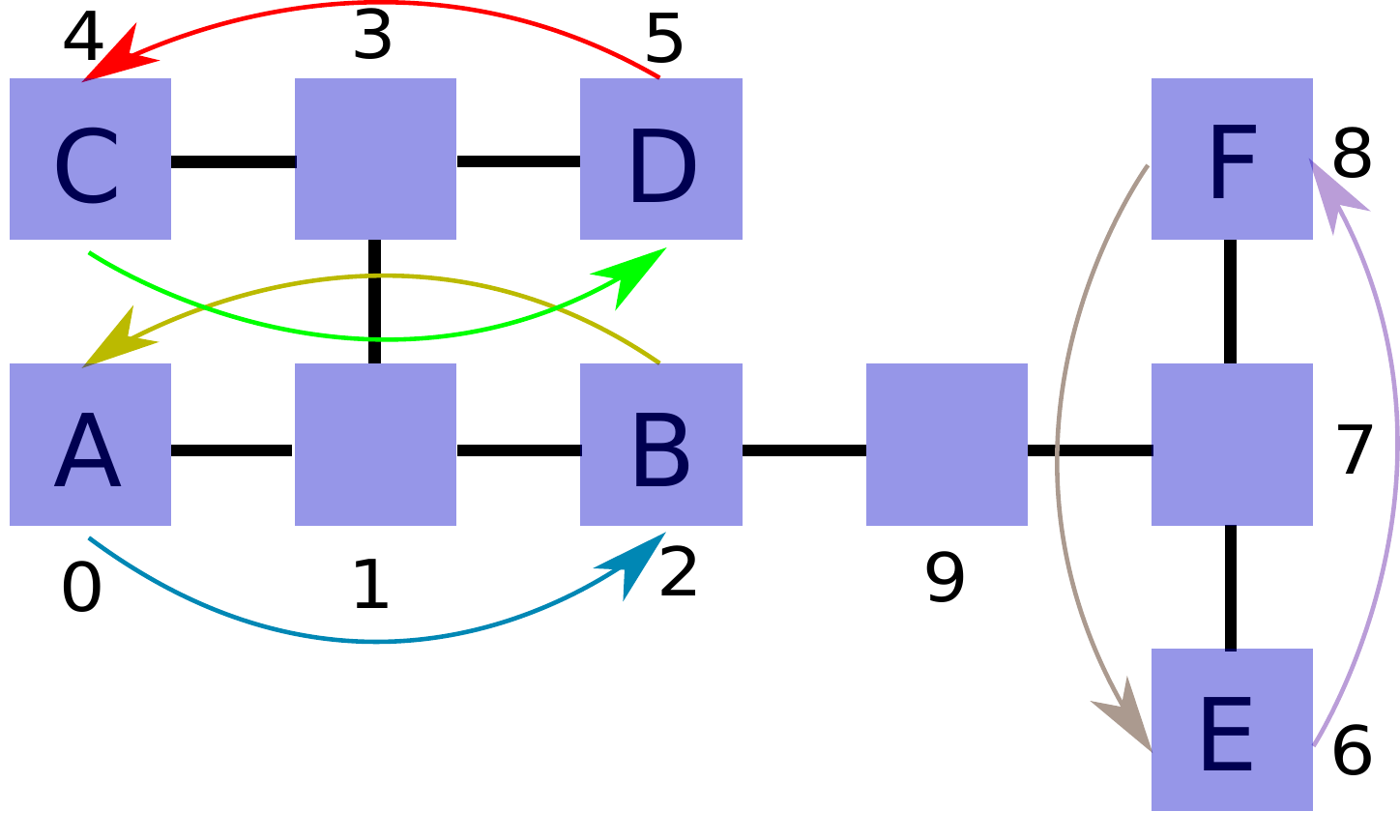}
            \caption{Second type of map expansion}
            \label{fig:mapgen_2}
    \end{subfigure}
    \caption{Map generation procedure.} 
    \label{fig:mapgen}
\end{figure}

The second set of maps was created specifically together with assignments so that the problems would be impossible to solve for the CARP algorithm. 
The maps and assignments were randomly generated by the following process:

\begin{enumerate}
\item Create a basic problem that is impossible to solve for the CARP algorithm depicted in Fig.~\ref{fig:mapgen_base}.
Arrows indicate the starting and goal positions of robots $A$ and $B$ on the graph.
CARP fails because the agents need to swap their positions while having the same distance to the only node they can use to avoid each other. Because CARP plans agents sequentially one by one while ignoring the subsequent agents, no ordering of these agents can solve this issue.
\item Pick random node that has only one edge associated with it.
\item Either add 2 nodes $A$ and $B$ to either side of this node if possible along with corresponding assignment of 2 agents -- The first agent going from $A$ to $B$ and the second one from $B$ to $A$.
The example of this step can be seen in Fig.~\ref{fig:mapgen_1}.
The alternative method to expand the map is to connect the same structure to it as in the Step 1 together with the same type of assignment, the example of which can be seen in Fig.~\ref{fig:mapgen_2}.
\item Repeat Steps 2 and 3 until the map of a required size is generated.
\end{enumerate}

\begin{figure}[htb]
    \centering
    \includegraphics[width=0.6\columnwidth]{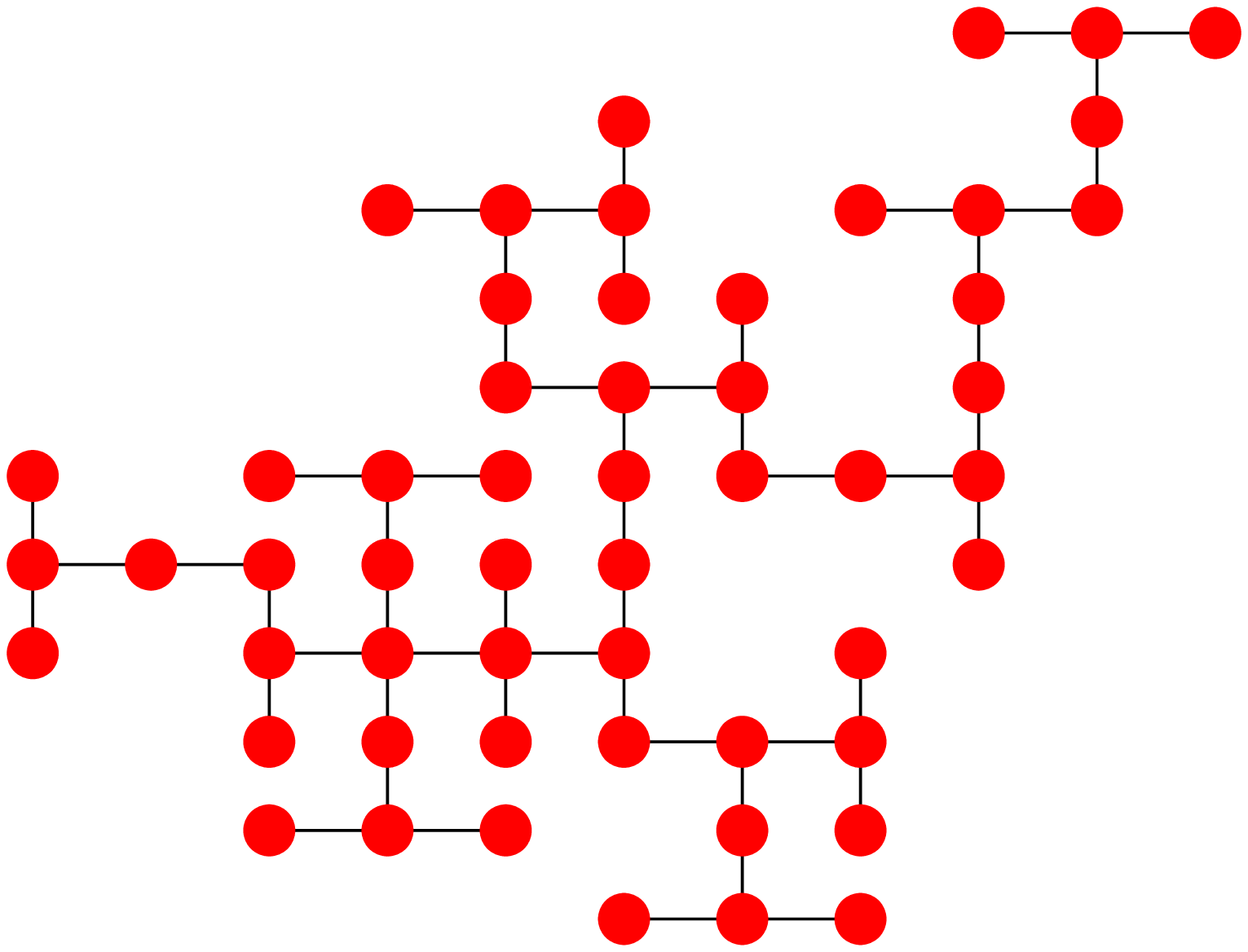}
    \caption{Example of a generated map.}
    \label{map_example}
\end{figure}

The example of a fully generated map following the previous steps can be seen in Fig.~\ref{map_example}.

\begin{figure*}[htb]
	

\begin{tabular}{ccc}
\includegraphics[width=0.33\columnwidth]{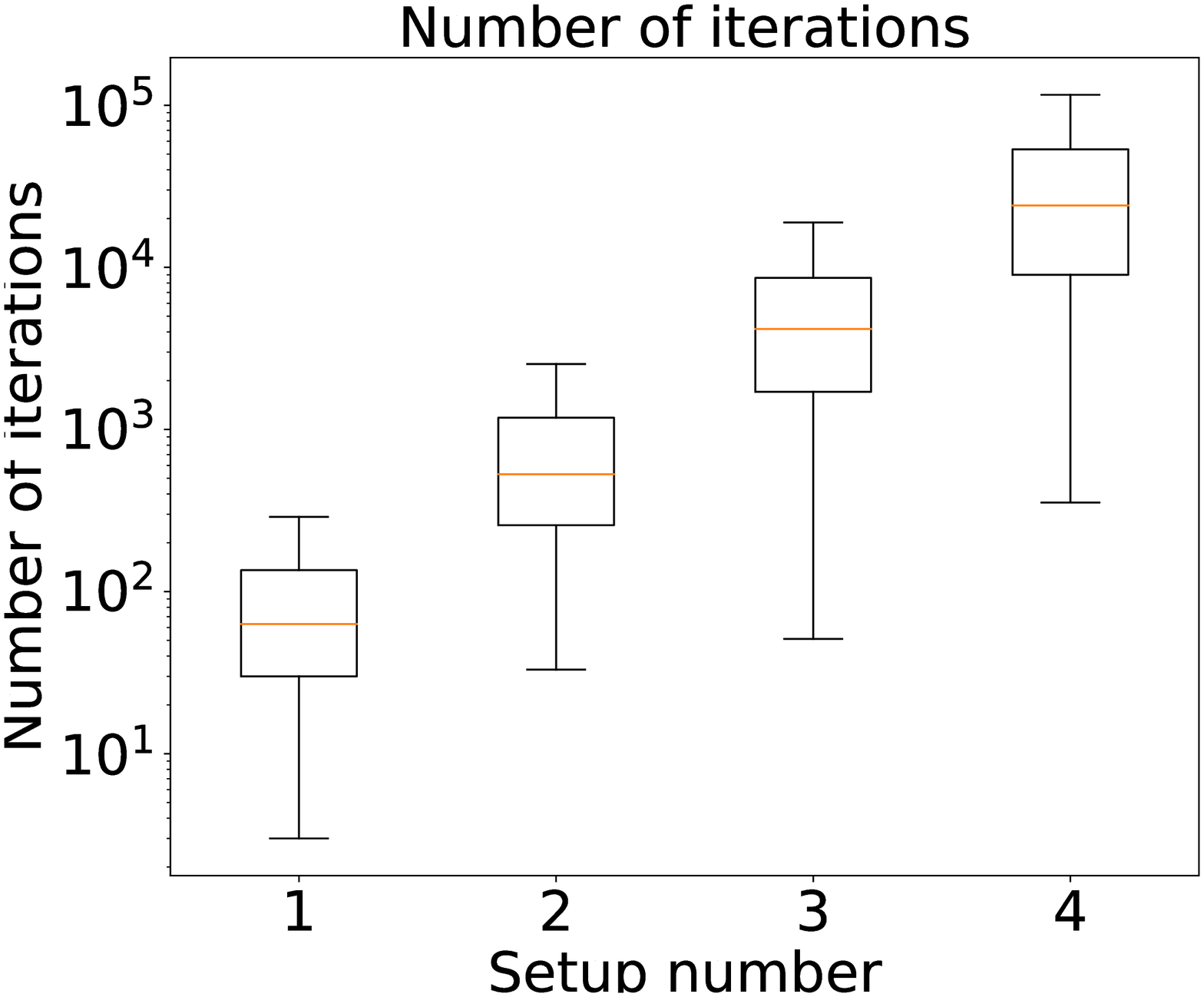} & \includegraphics[width=0.33\columnwidth]{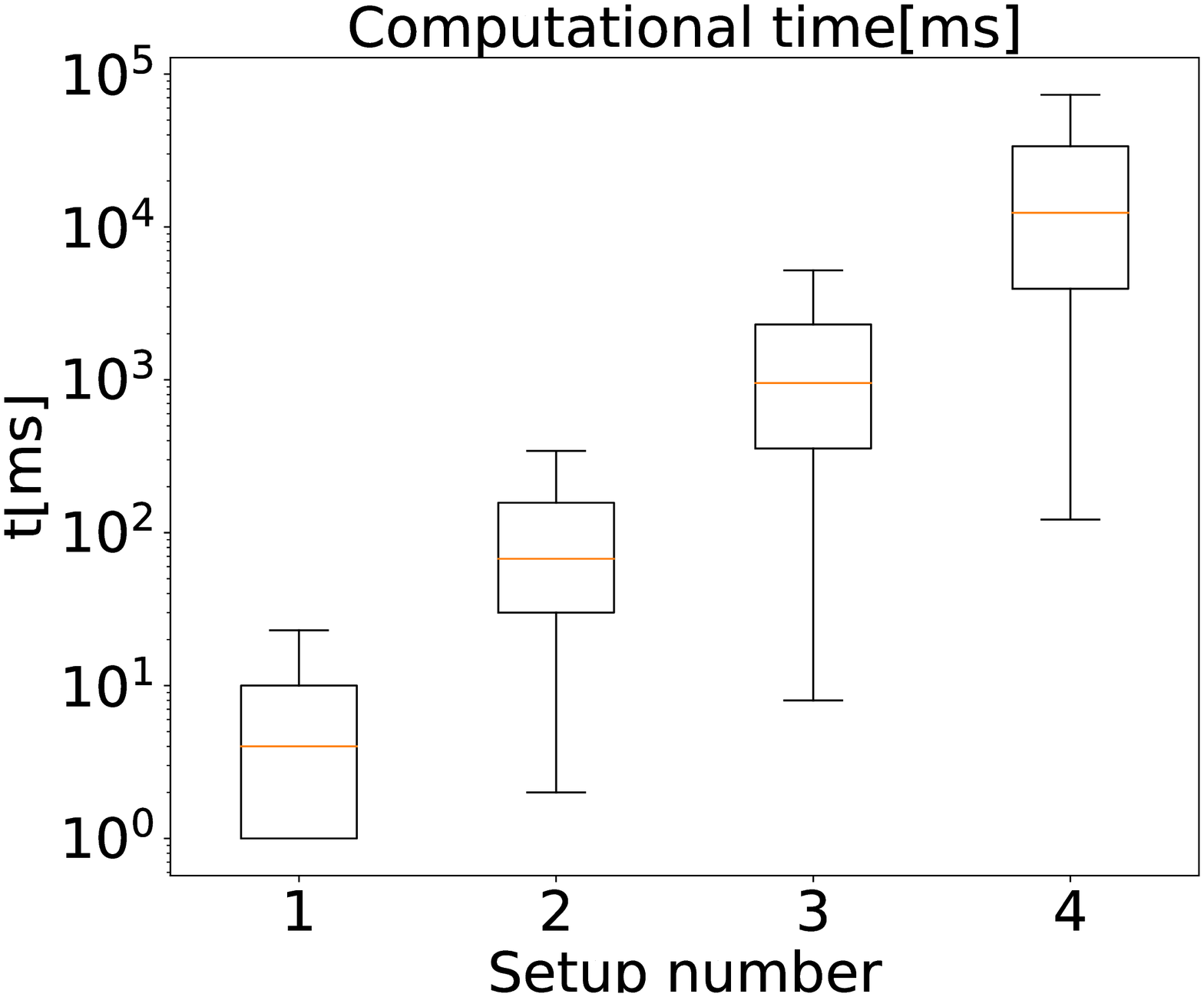} 
\includegraphics[width=0.33\columnwidth]{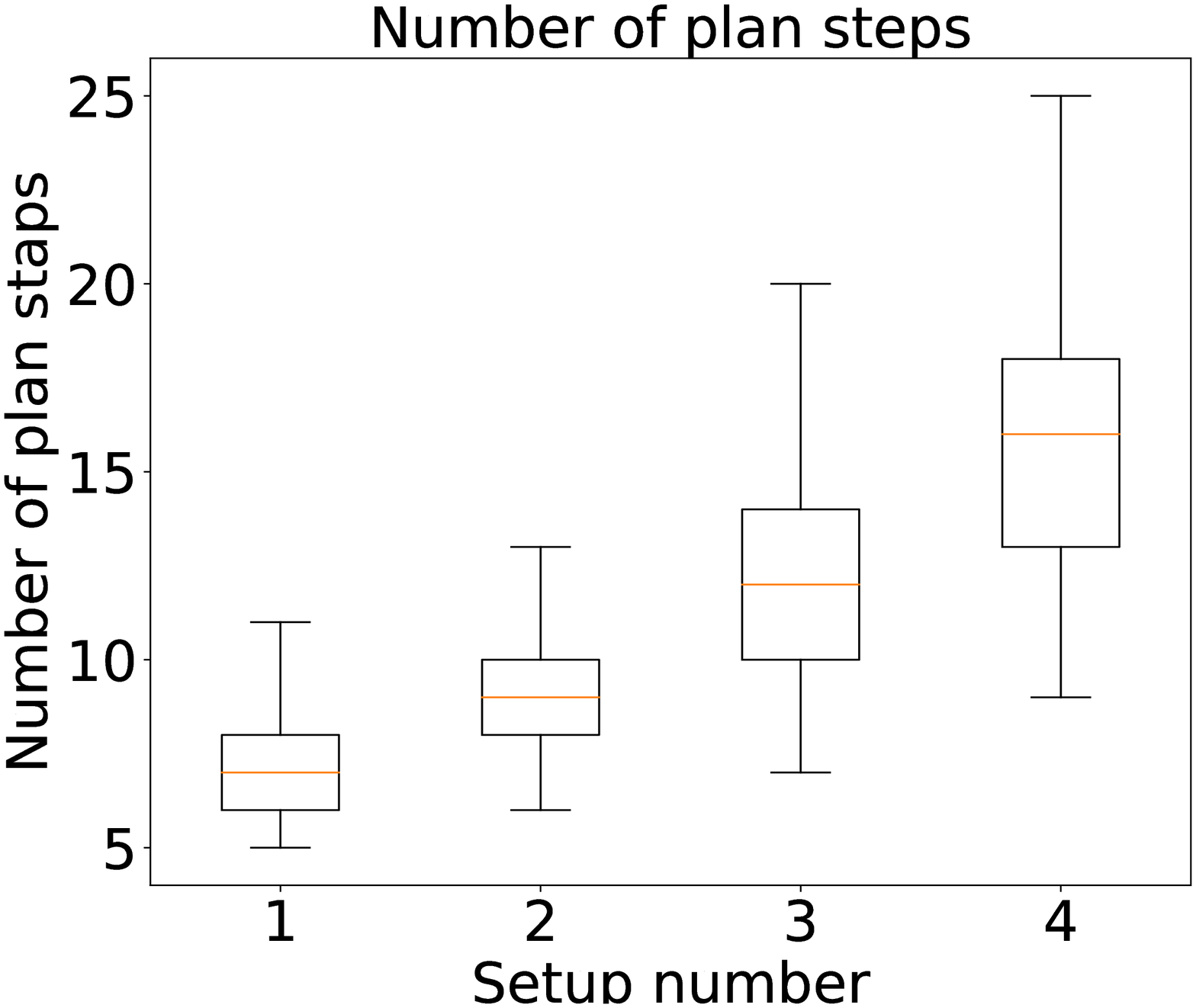} \\
\end{tabular}
\caption{Results of the proposed approach on assignments which CARP is unable to solve\cite{Hvezda2018ICINCO}.}
\label{fig:second_experiment}
\end{figure*}

The second set of experiments was carried out on the second set of maps with the aim to illustrate the behavior of the proposed algorithm on assignments that CARP algorithm can not solve.
The total of 400 different combinations of a map and assignment were generated: 100 each for 10, 20, 30 and 40 agents.
The results of this experiment can be seen in Fig.~\ref{fig:second_experiment}.
The setup numbers 1 to 4 correspond to the number of agents 10 to 40 respectively. 

For up to 30 agents the success rate is 100\% while it is decreased to 95\% for 40 robots.
Regarding the computational time results, the algorithm takes approximately 1 second to calculate the paths for each agent in assignments that are impossible to solve for CARP algorithm for up to 30 agents even with relatively complicated assignments. 

 \onecolumn

\section{\uppercase{Conclusion}}
\label{sec:conclusion}



This paper presented a novel approach for multi-robot coordination on a graph, which is based on a discrete version of RRT for multiple robots (MRdRRT) and significantly improves its performance in several steps of the algorithm.
Two additional steps inspired by RRT* were also introduced into the algorithm and which improve the quality of solutions the algorithm provides.
Moreover, the results of several experiments were performed that show the behavior of the algorithm in several various scenarios in different settings.
The results show that the proposed approach can solve problems containing tens of robots in tens of seconds, which is a significant improvement upon the original version of MRdRRT which was able to solve problems with up to ten robots in tens of seconds.
Moreover, the results also show that this approach is able to solve problems that are unsolvable for the CARP algorithm which is one of the best algorithms to be used in practice nowadays.

The future work should be focused on reducing the number of iterations the algorithm requires to find a solution as well as reducing the computational complexity. 
These improvements could be achieved for example by reducing the dimensionality of the problem by first clustering the robots into groups and then planning for these groups separately, but taking their respective solutions as obstacles.

\section*{\uppercase{Acknowledgements}}

\noindent This work has been supported by the European Union's Horizon 2020 research and innovation programme under grant agreement No 688117, by the Technology Agency of the Czech Republic under the project no.~TE01020197 \enquote{Centre for Applied Cybernetics}, the project Rob4Ind4.0 CZ.02.1.01/0.0/0.0/15\_003/0000470 and the European Regional Development Fund. 
The work of Jakub Hv\v{e}zda was also supported by the Grant Agency of the Czech Technical University in Prague, grant No.~SGS18/206/OHK3/3T/37.

\bibliographystyle{splncs03}
{
\bibliography{main}}

\end{document}